# Anatomy of Neural Language Models

Majd SALEH, Institute of Research and Technology b<>com, France, Majd.SALEH@b-com.com

Stéphane PAQUELET, Institute of Research and Technology b<>com, France

The fields of generative AI and transfer learning have experienced remarkable advancements in recent years especially in the domain of Natural Language Processing (NLP). Transformers have been at the heart of these advancements where the cutting-edge transformer-based Language Models (LMs) have led to new state-of-the-art results in a wide spectrum of applications. While the number of research works involving neural LMs is exponentially increasing, their vast majority are high-level and far from self-contained. Consequently, a deep understanding of the literature in this area is a tough task especially in the absence of a unified mathematical framework explaining the main types of neural LMs. We address the aforementioned problem in this tutorial where the objective is to explain neural LMs in a detailed, simplified and unambiguous mathematical framework accompanied by clear graphical illustrations. Concrete examples on widely used models like BERT and GPT2 are explored. Finally, since transformers pretrained on language-modeling-like tasks have been widely adopted in computer vision and time series applications, we briefly explore some examples of such solutions in order to enable readers to understand how transformers work in the aforementioned domains and compare this use with the original one in NLP.[1]

## 1 INTRODUCTION

Since the inception of transformers in 2017 [58], neural language models have revolutionized the fields of generative AI and transfer learning, especially in the field of Natural Language Processing (NLP). The new state-of-the-art results of transformer models pretrained on language modeling objectives have strongly inspired the development of similar solutions in Computer Vision (CV) and Time Series (TS) application fields. In order to better organize the literature of these exponentially growing research areas, several surveys have been recently published. For instance, Lin et al. explored in their survey [29] a wide variety of transformer variants and provided a detailed taxonomy of these latter. Min et al. explored the recent advances in NLP obtained by pretraining large language models [39]. Other works focused on more specific aspects/applications of transformers. For example, Dong et al. provided a survey on natural language generation methods [13]. Contextualized embedding methods have been investigated by Ethayarajh in [17] where the geometry of embeddings generated by widely used models like BERT [12] and GPT2 [48] have been compared. The evolution of semantic similarity methods has been explored by Chandrasekaran et al. in their survey [5]. The use of transformer language models in information retrieval applications has been explored by Gruetzemacher et al. [19] while prompting methods in NLP were reviewed by Liu et al. in their interesting work [31]. In computer vision and time series fields, the use of transformers has been explored in [26] and [61], respectively. Finally, topics about the interpretability and the explainability of neural NLP have been discussed in [34] and [15], respectively.

---

[1] Source codes are publically available at https://github.com/b-com/AnatomyNLM. They include a Python implementation of the formulas explained in this tutorial including an implementation of the GPT2 language model from scratch (inference pass).

On the one hand, the great surveys explored above provide a high level explanation where lots of theoretical and architectural aspects of neural language models are encapsulated and treated as black boxes. On the other hand, the original research works describing neural language models, especially transformer-based ones, are also introduced in a high level modular way. Consequently, a deep understanding of the literature in this area is a challenging task.

The objective of this tutorial is to provide a self-contained comprehensive anatomy of neural language models in a strict mathematical framework accompanied by clear graphical illustrations. Despite the focus on transformer LMs in recent works, we provide a unified framework covering Feedforward Neural Network (FFNN) LMs, Recurrent Neural Network (RNN) LMs and transformer LMs in order to highlight the common aspects of these models as well as the differences between them. We believe this is important to enable readers to better understand the motivations underlying the considered architectures. Two questions about neural language models are addressed: 1) how they work and, 2) why they work. The former is comprehensively addressed in the proposed mathematical framework while the latter, being a challenging open question related to the AI explainability, is discussed based on intuitions and practical experience.

In order to validate our mathematical framework, we derived the formulas that give the total number of trainable parameters of the considered LMs as functions of their hyper parameters. These formulas have been implemented in Python and their results were compared to trainable parameters' counters of TensorFlow with examples on widely used models like BERT [12] and GPT2 [48]. In addition, we provided a simple from-scratch implementation of transformer inference-pass equations with an example implementing GPT2 [48]. Our implementation was compared to the one of KerasNLP. The aforementioned comparisons, showing identical results, confirm that the anatomy of LMs explored in this tutorial is accurate and complete.

The paper is organized as follows: in the remaining subsections of the introduction, we explain the strict notation adopted in this tutorial in order to simplify the understanding of the proposed mathematical framework. Then, we explain how figures are organized. In Section 2, we explain the concept of word vector representations including one-hot vectors and dense real-valued representations i.e. embeddings. In Section 3, we define the two main types of language modeling i.e. autoregressive (AR) and autoencoding (AE) language modeling or equivalently masked language modeling (MLM). We explore some important applications of LMs and we introduce the concept of transfer learning [44] and its LM-based applications. Section 4 is devoted to introduce a high level overview of autoregressive language modeling. Three types of neural structures are considered i.e. FFNN- , RNN- and transformer-based LMs with a modular description of their main components. The concept of tying input and output embeddings [46] is then introduced. Besides, we explore how AR LMs are trained before discussing the limits of their input sequence length and how the order of input tokens is taken into account in the three considered LMs. In Section 5, we provide a comprehensive mathematical and graphical description of AR LMs where all their internal operations are explained besides listing their complete sets of trainable parameters (hence the title "Anatomy of neural language models"). A concrete example on the anatomy of the GPT2 [48] language model is provided. Section 6 discusses advanced topics on transformers including the application of masked language modeling with BERT [12] where the complete anatomy of this latter is provided. Training objectives of transformer-based AR and AE language models are then defined and discussed. The advantages of transformers are explored and compared to other types of neural networks in terms of their inductive bias and their computational efficiency. Finally, since transformers are widely used today in computer vision (CV) and time series (TS) applications with minor modifications on their structure, the use of transformers in these two domains is explored with concrete examples. In CV, the vision transformer ViT [14] is explored focusing on how embeddings are prepared. In TS analysis applications, a brief review of transformer-based solutions is provided before exploring the Time Series Transformer (TST) [64] where we focus on the preparation of input embeddings. Before diving further into this tutorial, readers who are not familiar with NLP terminology are referred to the Appendix A.

## 1.1 Notation

In order to avoid any ambiguity between scalars, vectors, matrices, function names, sets, tuples and sequences, we use a strict notation as described in Table 1. Vectors are represented in boldface *italic* small letters while matrices are represented in boldface *italic* CAPITAL letters. Sets are represented in $\mathcal{S}cript$ *italic* CAPITAL letters like $\mathcal{V} = \{v_1, v_2, \ldots, v_{|\mathcal{V}|}\}$. Note that curly brackets are used to group the elements of a set while round brackets are used to group the elements of a tuple/sequence. When multiple words are used in function names like **layerNorm(.)** or **selfAttention(.)**, lower camel case (camel case starting with a small letter) is adopted to facilitate the readability.

Table 1, Notation

| Entity | Letter case | Bold | *Italic* | $\mathcal{S}cript$ | Greek | Examples | Brackets |
|---|---|---|---|---|---|---|---|
| Scalar | - | no | yes | no | - | $a, b, x_i, x_i^{(j)}, X_{i,j}$ | - |
| Vector | small | yes | yes | no | - | $\boldsymbol{x}, \boldsymbol{v}, \boldsymbol{\omega}, \boldsymbol{\lambda}$ | $\boldsymbol{x} = [x^{(1)}, x^{(2)}, \ldots, x^{(n)}]^\top$ |
| Matrix | CAPITAL | yes | yes | no | - | $\boldsymbol{X}, \boldsymbol{V}, \boldsymbol{\Omega}$ | $\boldsymbol{X} = [\boldsymbol{x}_1, \boldsymbol{x}_2, \ldots, \boldsymbol{x}_m]$ |
| Function: fixed meaning | - | no | no | no | - | $\tanh(.), \text{layerNorm}(.)$ | $\sigma(x) = 1/(1 + e^{-x})$ |
| Function: generic symbol | - | no | yes | no | - | $f(.), g(.), \mathcal{J}(.)$ | $f(\boldsymbol{x}) = \boldsymbol{E}\,\boldsymbol{x}$ |
| Set of parameters | CAPITAL | no | yes | no | yes | $\theta, \Phi, \Psi$ | $\theta = \{\boldsymbol{E}, \boldsymbol{W}, \boldsymbol{b}, \Phi\}$ |
| Set in general | CAPITAL | no | yes | yes | - | $\mathcal{V}, \mathcal{R}, \mathcal{S}, \mathcal{X}, \mathcal{C}$ | $\mathcal{V} = \{v_1, v_2, \ldots, v_{|\mathcal{V}|}\}$ |
| Tuple or Sequence | CAPITAL | no | yes | yes | no | $\mathcal{V}, \mathcal{R}, \mathcal{S}, \mathcal{X}, \mathcal{C}$ | $\mathcal{S} = (w_1, w_2, \ldots, w_n)$ |

Concerning the use of subscripts and superscripts, we respect the following rules:

- We use subscripts to denote the index of a vector as a column of a matrix. For instance, a matrix $\boldsymbol{X} \in \mathcal{R}^{d \times n}$ can be represented in terms of its column vectors $\boldsymbol{x}_i \in \mathcal{R}^d \ \forall\, i \in \{1, \ldots, n\}$ as $\boldsymbol{X} = [\boldsymbol{x}_1, \boldsymbol{x}_2, \ldots, \boldsymbol{x}_n]$. We also use subscripts to denote the index of a matrix as a part of a bigger entity e.g. a tensor or a concatenated matrix. In this case, the column vectors of such matrix $\boldsymbol{X}_i$ will be denoted as $\boldsymbol{x}_{i,j}$ instead of $\boldsymbol{x}_{i_j}$ to simplify the notation. More generally speaking, sub-subscript representation is always replaced by this comma-separated-indices convention.
- We use superscripts with brackets to denote:
  o the indices of vector components: in this case we use round brackets. For example, the components of a vector $\boldsymbol{x} \in \mathcal{R}^d$ are written as $x^{(i)} \ \forall\, i \in \{1, \ldots, d\}$.
  o the indices of layers in a multilayer neural network: in this case we use angle brackets. For example, e.g. $\boldsymbol{X}^{[2]} = [\boldsymbol{x}_1^{[2]}, \boldsymbol{x}_2^{[2]}, \ldots, \boldsymbol{x}_n^{[2]}]$ represents an intermediate or an output matrix of the 2nd layer of a neural network.
- To denote a range of indices in sequences, we use the notation $w_{1:n}$ as a compressed version of $(w_1, w_2, \ldots, w_n)$. Similarly, we use the notation $w_{\leq n}$ to denote a sequence $(w_1, w_2, \ldots, w_n)$ if $n \geq 1$ or an empty sequence otherwise.

A matrix $\boldsymbol{X} \in \mathcal{R}^{d \times n}$ ($d$ rows $\times\, n$ columns) is generally represented as follows:

$$\boldsymbol{X} = \begin{bmatrix} X_{1,1} & X_{1,2} & \cdots & X_{1,n} \\ X_{2,1} & X_{2,2} & & X_{1,n} \\ \vdots & & \ddots & \vdots \\ X_{d,1} & X_{d,2} & \cdots & X_{d,n} \end{bmatrix}$$

However, when a matrix is represented in terms of its column vectors like $X = [x_1, x_1, \ldots, x_n]$, its detailed representation is given by:

$$X = [x_1, x_1, \ldots, x_n] = \begin{bmatrix} x_1^{(1)} & x_2^{(1)} & \cdots & x_n^{(1)} \\ x_1^{(2)} & x_2^{(2)} & & x_n^{(2)} \\ \vdots & & \ddots & \vdots \\ x_1^{(d)} & x_2^{(d)} & \cdots & x_n^{(d)} \end{bmatrix}$$

where the real-valued scalar $x_j^{(i)} \; \forall \; (i,j) \in \{1, \ldots, d\} \times \{1, \ldots, n\}$ represents the entry in the $i^{\text{th}}$ row and $j^{\text{th}}$ column.

All vectors in this tutorial are column vectors unless otherwise stated. Thus, a vector $x \in \mathcal{R}^d$ is expressed in terms of its components as $x = [x^{(1)}, x^{(2)}, \ldots, x^{(d)}]^{\mathsf{T}}$ with $\mathsf{T}$ representing the transpose operator.

Let $x_1$ and $x_2$ be two vectors from $\mathcal{R}^{d_1}$ and $\mathcal{R}^{d_2}$, respectively. Then, the concatenation of these two vectors can be represented using the $\oplus$ operator as: $x_1 \oplus x_2 = \left[x_1^{(1)}, x_1^{(2)}, \ldots, x_1^{(d_1)}, x_2^{(1)}, x_2^{(2)}, \ldots, x_2^{(d_2)}\right]^{\mathsf{T}}$. We use the same notation for concatenating matrices. However, matrix concatenation might mean vertical or horizontal stacking. For example, the concatenation of two matrices $X_1 = [x_{1,1}, \ldots, x_{1,n}] \in \mathcal{R}^{d_1 \times n}$ and $X_2 = [x_{2,1}, \ldots, x_{2,n}] \in \mathcal{R}^{d_2 \times n}$ is a matrix $X \in \mathcal{R}^{(d_1+d_2) \times n}$ that can be denoted as $X = X_1 \oplus X_2 = \begin{bmatrix} X_1 \\ X_2 \end{bmatrix}$. If the dimensions are ambiguous, we will state explicitly whether the concatenation is applied horizontally or vertically.

Elementwise multiplication of two vectors $x_1 \in \mathcal{R}^d$ and $x_2 \in \mathcal{R}^d$ (Hadamard product) is represented by the $\odot$ operator:

$$x_1 \odot x_2 = \left[x_1^{(1)} x_2^{(1)}, x_1^{(2)} x_2^{(2)}, \ldots, x_1^{(d)} x_2^{(d)}\right]^{\mathsf{T}}$$

Unless otherwise stated, activation functions like **tanh** and **sigmoid** will be applied on vectors or matrices element-wise. For example, $\tanh(x) = [\tanh(x^{(1)}), \tanh(x^{(2)}), \ldots, \tanh(x^{(n)})]^{\mathsf{T}}$.

The majority of functions described in this tutorial are parametrized functions representing neural networks' components. Such functions are represented as $Y = f_\theta(X)$ where $f$ is the function name, $\theta$ is the set of its parameters, $X$ is the input argument and $Y$ represents the output. This explicit differentiation between function parameters and input arguments enables a clearer modular representation of neural networks' components.

## 1.2  Organization of figures

Unless otherwise stated, figures are read from bottom to up and/or from left to right. Most of figures describe components of neural networks (NNs) as in the example shown in Figure 1. In this case, the NN component is shown at the middle of the figure. At the left side, the dimensionality of the trainable parameters are listed. At the right side, the set of trainable parameters, as well as the function which packages the considered NN component, are presented.

Vectors are represented horizontally or vertically in the figures to best fit the available space. The direction of vectors in figures doesn't imply that they are column or row vectors.

## 2  WORD VECTOR REPRESENTATIONS

In order to feed sequences of words into neural language models, words should be represented as numeric vectors. In this section, we explain the simple one-hot vector representation and the more advanced real-valued vector representation of words i.e. embeddings.

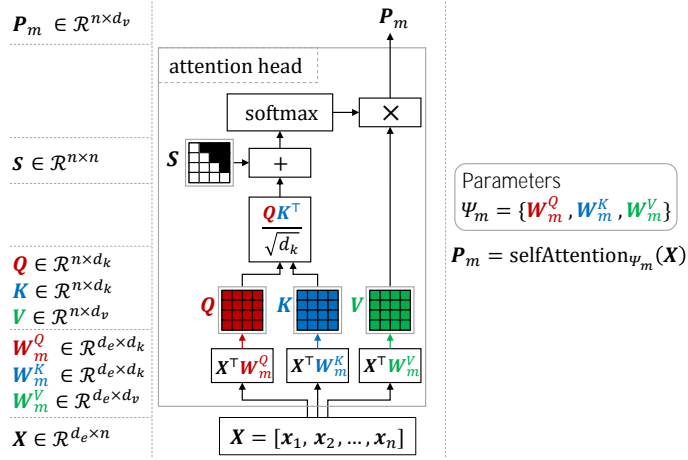

Figure 1. Organization of figures: an NN component is represented at the middle. At the left side, the dimensionality of the trainable parameters are provided. At the right side, the function packaging the NN component and the set of its parameters are presented.

### 2.1 One-hot vector representation

Each word in the vocabulary $\mathcal{V}$ can be represented by a one-hot vector: a vector of $|\mathcal{V}|$ components with only one component equals to one and the remaining components are zeros. The set of one-hot vectors will be denoted as $\mathcal{H} = \left\{ \boldsymbol{\omega} \in \{0,1\}^{|\mathcal{V}|} : \sum_{i=1}^{|\mathcal{V}|} \omega^{(i)} = 1 \right\}$. Table 2 shows some examples of this simple representation.

Table 2, One-hot vector representation.

| Index | Word | Notation | $|\mathcal{V}|$-components one-hot vector |
|---|---|---|---|
| 1 | ability | $v_1$ | $[1, 0, 0, 0, 0, \ldots, 0]^\top$ |
| 2 | above | $v_2$ | $[0, 1, 0, 0, 0, \ldots, 0]^\top$ |
| 3 | apple | $v_3$ | $[0, 0, 1, 0, 0, \ldots, 0]^\top$ |
| $\vdots$ | $\vdots$ | $\vdots$ | $\vdots$ |
| $|\mathcal{V}|$ | zoo | $v_{|\mathcal{V}|}$ | $[0, 0, 0, 0, 0, \ldots, 1]^\top$ |

### 2.2 Real-valued dense vector representations: embeddings

With the real-valued dense vector representation, each word in the vocabulary $\mathcal{V}$ is represented by a dense real-valued vector of $d_e$ components with $d_e \ll |\mathcal{V}|$. These vectors are called "embeddings". Linguistic relationships between words are expressed in their embeddings [5]. For example, similar words should have similar embeddings. This way, semantic similarity between words can be measured using, for example, the cosine or the Euclidean distance between the corresponding embeddings. For a visual illustration, the reader is referred to the online embedding projector of TensorFlow [56] where different pretrained embeddings can be visualized. The visualization is based on dimensionality reduction algorithms like Principal Component Analysis (PCA). The cosine distance and the Euclidian distance between a selected word and the remaining words in the vocabulary are listed beside the visualization panel.

Let $\boldsymbol{E} = \left[\boldsymbol{e}_1, \boldsymbol{e}_2, \ldots, \boldsymbol{e}_{|\mathcal{V}|}\right] \in \mathcal{R}^{d_e \times |\mathcal{V}|}$ denote the embedding matrix where each column $\boldsymbol{e}_i$ represents the embedding of one word in the vocabulary. Mapping one-hot vectors to embeddings is accomplished using $\mathcal{H} \to \mathcal{R}^{d_e}: \boldsymbol{\omega} \to \boldsymbol{E}\boldsymbol{\omega}$.

Note that multiplying $\boldsymbol{E}$ by a one-hot vector is equivalent to selecting the corresponding embedding. To show this, Table 3 and Figure 2 show a concrete example with a sequence of tokens $\mathcal{S} = (w_1, w_2, \ldots, w_9)$ that represents the sentence "He never said "better late than never"".

Table 3, Representing a sequence of tokens with vectors.

| $n = 9$ | He | never | said | " | better | late | than | never | " |
|---|---|---|---|---|---|---|---|---|---|
| $\mathcal{S} = (w_1, w_2, \ldots, w_9)$ | $w_1$ | $w_2$ | $w_3$ | $w_4$ | $w_5$ | $w_6$ | $w_7$ | $w_8$ | $w_9$ |
| $\mathcal{V}$ | $v_{1344}$ | $v_{23500}$ | $v_{9003}$ | $v_6$ | $v_{3250}$ | $v_{713}$ | $v_{89}$ | $v_{23500}$ | $v_7$ |
| One-hot: $\boldsymbol{\omega}_i \in \mathcal{H}$ | $\boldsymbol{\omega}_1$ | $\boldsymbol{\omega}_2$ | $\boldsymbol{\omega}_3$ | $\boldsymbol{\omega}_4$ | $\boldsymbol{\omega}_5$ | $\boldsymbol{\omega}_6$ | $\boldsymbol{\omega}_7$ | $\boldsymbol{\omega}_8$ | $\boldsymbol{\omega}_9$ |
| Embeddings: $\boldsymbol{x}_i \in \mathcal{R}^{d_e}$ | $\boldsymbol{x}_1$ | $\boldsymbol{x}_2$ | $\boldsymbol{x}_3$ | $\boldsymbol{x}_4$ | $\boldsymbol{x}_5$ | $\boldsymbol{x}_6$ | $\boldsymbol{x}_7$ | $\boldsymbol{x}_8$ | $\boldsymbol{x}_9$ |

Let $\boldsymbol{\Omega} = [\boldsymbol{\omega}_1, \boldsymbol{\omega}_2, \ldots, \boldsymbol{\omega}_9]$ denote the matrix representation of the considered sequence in terms of the corresponding one-hot vectors. Then, the dense representation matrix of the sequence $\boldsymbol{X} = [\boldsymbol{x}_1, \boldsymbol{x}_2, \ldots, \boldsymbol{x}_9]$ (each column represents a token) is obtained using $\boldsymbol{X} = \boldsymbol{E}\,\boldsymbol{\Omega}$ as shown in Figure 2-b.

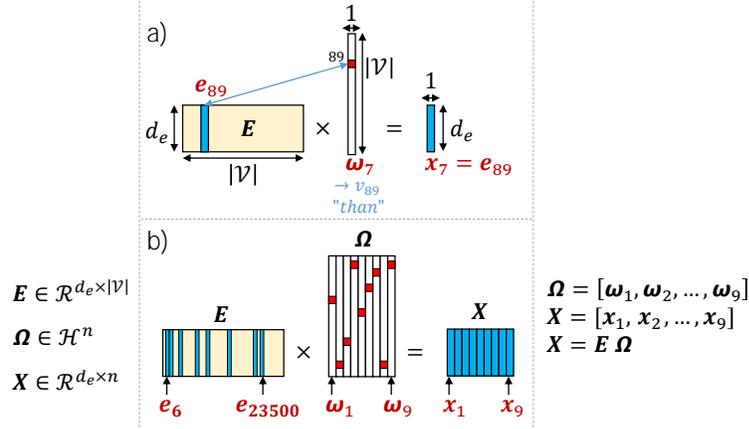

Figure 2. a) extracting the embedding of the word $\boldsymbol{v}_{89}$ "than". b) extracting the embeddings of the sequence given in Table 3. This Figure is inspired by [25]

Several approaches can be used to generate embeddings ranging from simple methods based on word co-occurrence matrix to more complex methods based on machine learning [5]. These latter use large text corpora to train classifiers on specific tasks that enable capturing the semantic similarity between words. Such training tasks are generally based on the distributional hypothesis [20]. This hypothesis, established by linguists, states roughly that the meaning of a word can be defined by its distribution in language use [25]. In other words, similar words occur together frequently [5]. Based on this hypothesis, embeddings are built using measures of words' co-occurrence in the training corpus. Two main families of methods can be distinguished: 1) static embedding methods and, 2) dynamic or contextualized embedding methods. With the first family, each word in the vocabulary has a single fixed embedding [17]. The most widely used pre-trained static embeddings are word2vec [36, 37], GloVe [45] and fastText [3]. With the second family, the embedding of a word depends on its context[2]. Embeddings generated using transformer-based models like BERT [12] and GPT [4, 43, 47, 48] are instances of this contextualized embedding family. Methods for contextualized embeddings will be explored as an integrated part of transformer-based language models while methods for pre-training static embeddings fall out of the scope of this tutorial.

---

[2] Training embeddings, both static and contextualized ones, exploits the context of the word being represented. The name "contextualized" comes from the way how words are represented in post-training use. Contextualized embedding methods will generate a new representation for the same word once its context changes.

# 3 LANGUAGE MODELING

## 3.1 Language modeling types

There are two main types of language modeling (LM): 1) autoregressive (AR) language modeling or equivalently causal language modeling where the objective is to predict upcoming tokens from prior token context, and 2) autoencoding (AE) language modeling or equivalently Masked Language Modeling (MLM) where the objective is to predict some masked token(s) given the remaining tokens in the sequence from both directions.

AR LMs represent the standard family which has been studied from more than one century. They evolved from n-gram language models [28, 35, 51] to neural language models. These latter started by simple Feed-Forward Neural Networks (FFNNs)-based LMs [2], evolved to Recurrent Neural Networks (RNNs)-based LMs [18, 38, 46, 53, 63] and recently evolved to the cutting edge transformer-based LMs [4, 43, 47, 48, 58]. On the other hand, AE language modeling was not straightforwardly applicable with FFNN and RNN architectures. Thanks to the transformer architecture, AE language modeling was successfully applied using bidirectional transformer encoders [12, 62].

## 3.2 Applications of language modeling

Language modeling is of great importance in NLP for two main reasons: 1) it represents a principal component in several NLP systems. For instance, AR language modeling is used in machine translation, abstractive summarization, speech recognition, optical character recognition (OCR), and other applications that involve generating texts. Another example is the important role of AE language modeling in sequence labeling e.g. part of speech (POS) tagging and named entity recognition (NER), extractive summarization, and other applications where bidirectional access to the context is authorized. 2) Language modeling is a principal pretraining objective that enables training large language models and making them exploitable in a wide range of applications via transfer learning as will be explained in Section 3.5.

## 3.3 Autoregressive language modeling

Let $\mathcal{C} = (c_1, c_2, \ldots, c_T)$ denote a corpus of $T$ tokens (in practice, $T$ might be several billions of tokens). AR neural language modeling aims at developing a model that maximizes the probability of the entire corpus $p(\mathcal{C})$. In practice, this problem is formulated as a minimization of the following negative log likelihood [47]:

$$\mathcal{J}_{AR}(\mathcal{C}) = -\sum_{i=0}^{T-1} \log p_\theta(c_{i+1}|c_{i-n+1:i}) \quad (1)$$

where the following considerations are employed:
1- the chain rule of probability is used to factorize the joint probability $p(\mathcal{C})$ as $p(\mathcal{C}) = \prod_{i=0}^{T-1} p(c_{i+1}|c_{\leq i})$.
2- the context window size is limited to $n$ tokens due to implementation restrictions.
3- the probability distribution $p$ is modeled using a neural network whose trainable parameters are $\theta$. Thus, the modeled probability is denoted as $p_\theta$; the estimated neural approximation of $p$.

Equation 1 will be discussed in more detail in Section 4.6 where the cross entropy loss will be introduced. The practical limits of the context window size will be discussed in Section 4.7 after exploring the main types of neural LMs.

Let $\mathcal{S}_{in} = (w_1, w_2, \ldots, w_n)$ denote a context sequence of size $n$, $w_{n+1}$ denote the token to be predicted and $\mathcal{S} = (w_1, w_2, \ldots, w_n, w_{n+1})$ denote the sequence that contains both $\mathcal{S}_{in}$ and $w_{n+1}$. AR language models are trained iteratively to maximize the probability of training sequences $p(\mathcal{S}) = \prod_{i=0}^{n} p(w_{i+1}|w_{\leq i})$.

### 3.4 Masked Language Modeling (MLM)

Given a sequence of tokens $S = (w_1, w_2, ..., w_n)$, the objective of masked language modeling is to predict some masked token(s) given the remaining tokens in the sequence [12]. Let $S_c$ denote a corrupted version of $S$ where some randomly selected tokens (say 15% of $S$ tokens) are masked i.e. each of them is replaced by a special token [MASK]. Let $S_m$ represent the masked tokens. MLM aims at maximizing the probability $p(S_m|S_c)$.

For example, if $S = (\text{I, knew, it, was, going, to, rain, but, I, forgot, to, take, my, umbrella})$, then a corrupted version of the sequence could be $S_c = (\text{I, knew, it, [MASK], [MASK], to, rain, but, I, [MASK], to, take, my, umbrella})$ and the corresponding masked tokens are $S_m = (\text{was, going, forgot})$. MLM will be explained in detail in Section 6.1.

### 3.5 Transfer learning

In machine learning, transfer learning (or inductive transfer) is a technique that aims at training a model to acquire knowledge from a source domain $\mathcal{D}_S$ using a generic training task $\mathcal{T}_S$ such that this model can be used to improve the learning of the target predictive function $f_T(.)$ trained on a different but related task $\mathcal{T}_T$ in the target domain $\mathcal{D}_T$ [44].

In NLP, several optimization objectives were investigated for generating text representations that are useful for transfer. For instance, language modeling, machine translation and discourse coherence were employed as pretraining objectives [47]. Recently, language modeling as pretraining objective showed very impressive transfer results with models like GPT of OpenAI [4, 43, 47, 48] and BERT of Google [12], to cite a few. Multiple training objectives can be used together when pretraining a model. For instance, BERT is pretrained by minimizing the sum of masked language modeling loss and Next Sentence Prediction (NSP) loss as will be explained in Sections 6.1 and 6.2.

Pretraining LMs aims at training models how to represent human languages. It is usually performed using unlabeled data in a self-supervised learning framework which enables exploiting huge amounts of natural language corpora. The pretrained models can be used in a wide spectrum of downstream tasks in two ways [12]: 1) pretrained models are used as feature extractors: they are used to represent textual data in real-valued dense vectors (features) that can be used to solve the target NLP tasks, and 2) pretrained models are fine-tuned on the downstream task using a relatively small labeled corpus. In the fine-tuning strategy, training implies adjusting all the weights of the pretrained model to perform better on the downstream task while in the feature extraction strategy, the pretrained model weights are frozen. Note that fine-tuning can be also applied on a portion of the pretrained model's weights where the weights of the base layers can be frozen and the weights of higher layers can be set trainable.

## 4 HIGH LEVEL OVERVIEW OF AUTOREGRESSIVE LANGUAGE MODELING

In this section, a high level explanation of different types of AR LMs is provided. LMs based on FFNNs, RNNs and transformers are considered. The objective is to show the major common aspects of these LMs as well as the major architectural differences between them. Thus, the three LM types will be presented in a modular way i.e. focusing on the input and the output of their main components, while their detailed architectures will be presented in later sections.

### 4.1 Input and output of language models

Given a context represented by a sequence of tokens $S = (w_1, w_2, ..., w_n)$, the objective of AR language modelling is to predict the next token $w_{n+1}$. Neural networks are used to model the conditional probability $p(w_{n+1}|w_{1:n})$. They are trained i.e. their trainable parameters $\theta$ are adjusted, such that the modelled probability distribution $p_\theta(w_{n+1}|w_{1:n})$ is as close as possible to the true one. Figure 3 shows the input and output of autoregressive language models. The input sequence $S = (w_1, w_2, ..., w_n)$ is first converted into the corresponding one-hot vectors $\boldsymbol{\omega}_i \in \mathcal{H} \ \forall \ i \ \in \{1, ..., n\}$ where $\mathcal{H} = \{\boldsymbol{\omega} \in \{0,1\}^{|\mathcal{V}|} : \sum_{i=1}^{|\mathcal{V}|} \omega^{(i)} = 1\}$ with $|\mathcal{V}|$ representing the vocabulary size. An AR LM takes this input and generates a vector $\hat{\boldsymbol{y}} \in \mathcal{R}^{|\mathcal{V}|}$ that represents a probability distribution $p_\theta$ over the vocabulary. In the example shown in Figure 3, the context is "I knew it was going to rain but I forgot to take my" while $\hat{\boldsymbol{y}}$ is shown in hot color map indicating that the word "umbrella" corresponds to the highest probability compared to the remaining words in the vocabulary $\mathcal{V}$.

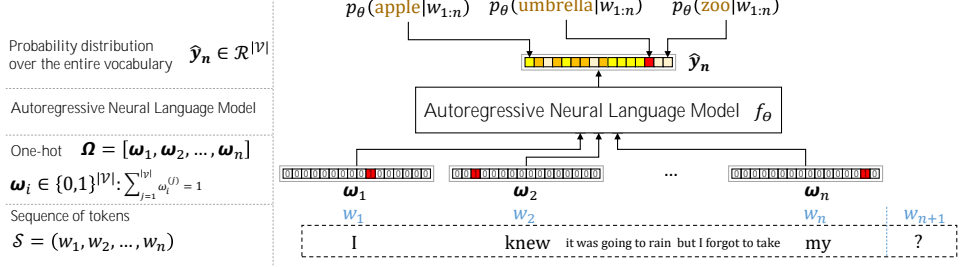

Figure 3. Input and output of language models

Let us start exploring language models based on FFNNs, transformers and RNNs, respectively[3]. We assume that each of these networks consists of $L$ layers. We use $l \in \{1, ..., L\}$ to express the layer index while the index $0$ expresses the input.

## 4.2 Modules of FFNN-based Language Models

Figure 4 shows the use of a FFNN for modeling the probability distribution $P_\theta$. The first step is to represent input tokens by dense real-valued vectors i.e. embeddings. This can be achieved using an embedding matrix $\boldsymbol{E} \in \mathcal{R}^{d_0 \times |\mathcal{V}|}$ with $d_0 \ll |\mathcal{V}|$. The matrix $\boldsymbol{E}$ represents a trainable parameter[4] while $d_0$ represents a hyper parameter that can be set empirically. We will call $\boldsymbol{E}$ the input embedding matrix. Selecting the embeddings of input tokens that correspond to the one-hot vectors $\boldsymbol{\omega}_i$ is performed using equation 2:

$$\boldsymbol{x}_i = \boldsymbol{E}\boldsymbol{\omega}_i \quad \forall i \in \{1, ..., n\} \tag{2}$$

The embeddings of the input tokens $\boldsymbol{x}_i \in \mathcal{R}^{d_0} \forall i \in \{1, ..., n\}$ are concatenated in one vector $\boldsymbol{h}_n^{[0]} \in \mathcal{R}^{nd_0}$ which is then fed to a feedforward neural network $\text{ffnn}_\Phi$ whose trainable parameters are denoted as $\Phi$. This FFNN consists of $L$ dense layers (fully connected layers). The output of this network $\boldsymbol{h}_n^{[L]} \in \mathcal{R}^{d_L}$ is expressed in equation 3:

$$\boldsymbol{h}_n^{[L]} = \text{ffnn}_\Phi\left(\boldsymbol{h}_n^{[0]}\right) \tag{3}$$

The vector $\boldsymbol{h}_n^{[L]}$ is then multiplied by a trainable matrix $\boldsymbol{U} \in \mathcal{R}^{|\mathcal{V}| \times d_L}$, called the output embedding matrix, in order to generate the logits $\boldsymbol{z}_n \in \mathcal{R}^{|\mathcal{V}|}$ as in equation 4:

$$\boldsymbol{z}_n = \boldsymbol{U}\boldsymbol{h}_n^{[L]} \tag{4}$$

To convert the logits into a probability distribution $\hat{\boldsymbol{y}} \in \mathcal{R}^{|\mathcal{V}|}$, the softmax function is used:

$$\hat{\boldsymbol{y}}_n = \text{softmax}(\boldsymbol{z}_n) = \left[\frac{\exp(z_n^{(1)})}{\sum_{j=1}^{|\mathcal{V}|} \exp(z_n^{(j)})}, \frac{\exp(z_n^{(2)})}{\sum_{j=1}^{|\mathcal{V}|} \exp(z_n^{(j)})}, \ldots, \frac{\exp(z_n^{(|\mathcal{V}|)})}{\sum_{j=1}^{|\mathcal{V}|} \exp(z_n^{(j)})}\right] \tag{5}$$

Thus, this language model is a simple neural network that generates a probability distribution over the vocabulary starting from the one-hot representations of the input tokens $\boldsymbol{\Omega} = [\boldsymbol{\omega}_1, \boldsymbol{\omega}_2, ..., \boldsymbol{\omega}_n]$. This can be expressed as follows:

$$\hat{\boldsymbol{y}}_n = f_\theta(\boldsymbol{\Omega}) \tag{6}$$

---

[3] FFNN, transformer and RNN LMs are not presented in their historical order i.e. FFNN, RNN, then transformer. We present FFNN LMs first because they are the simplest. Then we explore transformers because they have more similar aspects with FFNN LMs than the RNN LMs. Finally, we present RNN LMs.

[4] Pretrained embeddings can be used instead of training the embedding matrix from scratch.

where $\theta = \{\Phi, E, U\}$ represents the set of trainable parameters. From this probability distribution $\hat{y}_n$, the model proposes the word of the highest probability as the predicted next word.

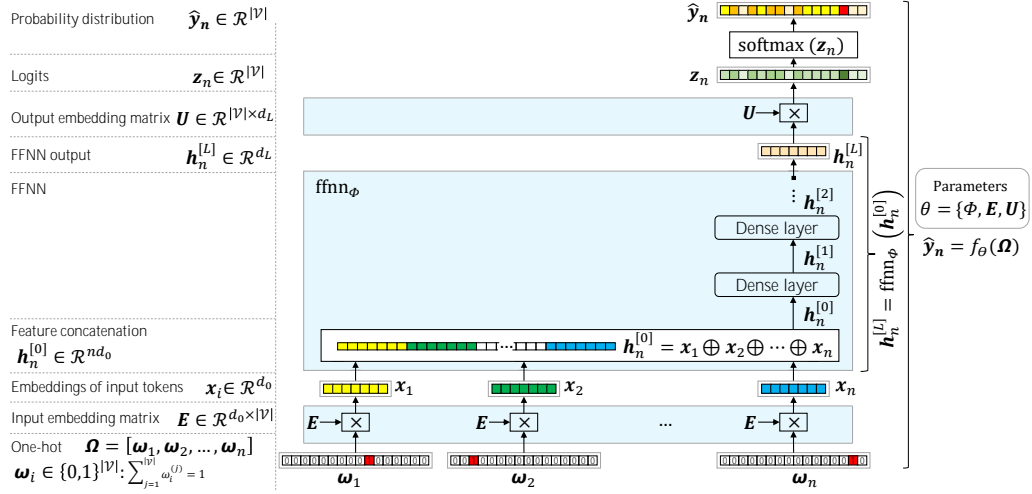

Figure 4. FFNN language model

Note that the index $n$ might have been dropped from $\hat{y}_n$, $z_n$ and $h_n^{[l]}$ without ambiguity. However, we preferred to keep it for two reasons: 1) to facilitate the comparison with other types of language models, and 2) to emphasize the fact that a fixed width sliding window is used to provide the context.

### 4.3 Modules of Transformer-based Language Models

A transformer-based language model is depicted in Figure 5. As with FFNN LMs, the first step is to pass from one-hot representation of input tokens to dense embeddings as shown in equation 2. Then, contrary to FFNNs where embeddings are concatenated, the embedding vector of each token is fed to the transformer which generates modified versions of these embeddings. In fact, a transformer generates these modified embeddings after performing a systematic analysis of the relationships between input tokens. This analysis is achieved by the self-attention mechanism that will be explained in detail in Section 5.2. The outputs of the transformer are called contextualized embeddings or dynamic embeddings. Each of them is based both on the initial embedding of a token and on its context. The way how transformers process input sequences doesn't natively take into account the order of tokens. To this end, transformers add an explicit positional encoding $\lambda_i \in \mathcal{R}^{d_0}$ to the input embeddings $x_i \in \mathcal{R}^{d_0}$. These positional encodings aim at representing the position of the token regardless what the token is. They might be trained from scratch or, alternatively, fixed sinusoidal formulas can be used as in [58]. In this section, we consider positional encodings as trainable parameters. Let $\Lambda = [\lambda_1, \lambda_2, \ldots, \lambda_n] \in \mathcal{R}^{d_0 \times n}$ denote the positional encoding matrix, $X = [x_1, x_2, \ldots, x_n] \in \mathcal{R}^{d_0 \times n}$ represent the embedding of input tokens ($X = E\Omega$), then, the input of the transformer $H^{[0]} = \left[h_1^{[0]}, h_2^{[0]}, \ldots, h_n^{[0]}\right]$ is given as:

$$H^{[0]} = \Lambda + X \tag{7}$$

The transformer can be then encapsulated as a parametrized function:

$$H^{[L]} = \text{transformer}_\Phi(H^{[0]}) \tag{8}$$

where $H^{[L]} = \left[h_1^{[L]}, h_2^{[L]}, \ldots, h_n^{[L]}\right] \in \mathcal{R}^{d_L \times n}$ represents its output and $\Phi$ represents its trainable parameters.

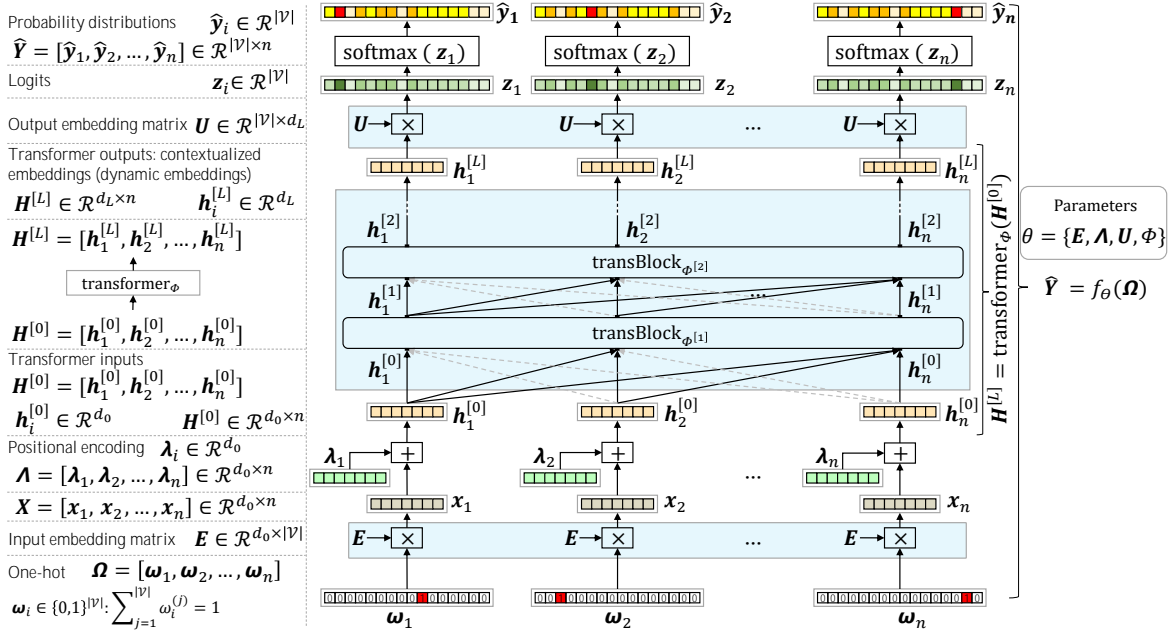

Figure 5. Transformer language model

As in FFNN LMs, the outputs of the transformer are multiplied by the output embedding matrix $U \in \mathcal{R}^{|\mathcal{V}| \times d_L}$ to produce the logits $z_i \in \mathcal{R}^{|\mathcal{V}|}$:

$$z_i = U h_i^{[L]} \quad \forall i \in \{1, \dots, n\} \tag{9}$$

The logits are then projected into probability distributions using the softmax function:

$$\hat{y}_i = \text{softmax}(z_i) = \left[ \frac{\exp(z_i^{(1)})}{\sum_{j=1}^{|\mathcal{V}|} \exp(z_i^{(j)})}, \frac{\exp(z_i^{(2)})}{\sum_{j=1}^{|\mathcal{V}|} \exp(z_i^{(j)})}, \dots, \frac{\exp(z_i^{(|\mathcal{V}|)})}{\sum_{j=1}^{|\mathcal{V}|} \exp(z_i^{(j)})} \right] \quad \forall i \in \{1, \dots, n\} \tag{10}$$

Note that, contrary to the FFNN LMs, transformers provide an estimated probability distribution $\hat{y}_i$ corresponding to each input token $\omega_i$, not only for the last one. Despite that we need only the last one to predict the next word given the context $\omega_{1:n}$, the other distributions are employed in the training phase where all of them are compared to the true probability distributions as will be explained in detail in Section 4.6.

Finally, a transformer-based language model can be expressed as:

$$\hat{Y} = f_\theta(\Omega) \tag{11}$$

where $\theta = \{E, \Lambda, U, \Phi\}$ represent its trainable parameters. As will be explained in Section 5.2, the function $f_\theta$ in AR LMs utilizes a binary mask to prevent accessing the future tokens i.e. to make $\hat{y}_i$ dependent on $\omega_{1:i}$ only.

### 4.4 Modules of RNN-based Language Models

An RNN layer is shown in Figure 6-a. The output of the layer $l$ at the time step $i$, $h_i^{[l]}$, is dependent on the output of the previous layer at the same time step $h_i^{[l-1]}$ as well as the output of the same layer at the previous time step $h_{i-1}^{[l]}$. Thus, $h_i^{[l]}$ is given in equation 12:

$$h_i^{[l]} = \text{rnn}_{\Phi^{[l]}}\left(h_{i-1}^{[l]}, h_i^{[l-1]}\right) \tag{12}$$

where $\Phi^{[l]}$ represents the trainable parameters of the layer $l$. To facilitate the comparison between RNN LMs and the other considered LMs, we will represent RNNs unrolled in time as shown in Figure 6-b.

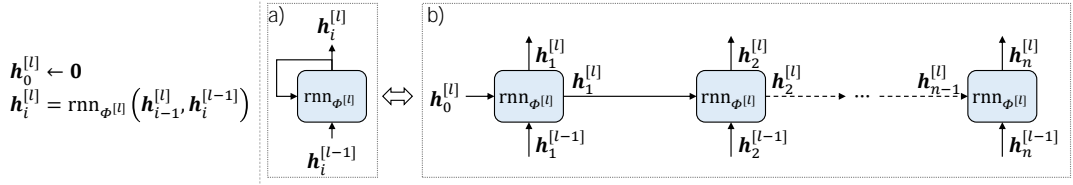

Figure 6. a) RNN layer, b) RNN layer unrolled in time

Figure 7 shows a recurrent neural network as a language model. The order of input tokens is implicitly taken into account in the recurrence mechanism. As with FFNN and transformer LMs, the first step is to select the embeddings of the input tokens from the input embedding matrix $E$. This is achieved as shown in equation 2.

The resultant embeddings of input tokens constitute the input of the RNN $h_i^{[0]} \leftarrow x_i = E\,\omega_i$. The corresponding outputs of the RNN, $h_i^{[L]} \in \mathcal{R}^{d_L}$, are then obtained using equation 12 with vectors $h_0^{[l]}$ initialized by zeroes $\forall l \in \{1, \dots, L\}$.

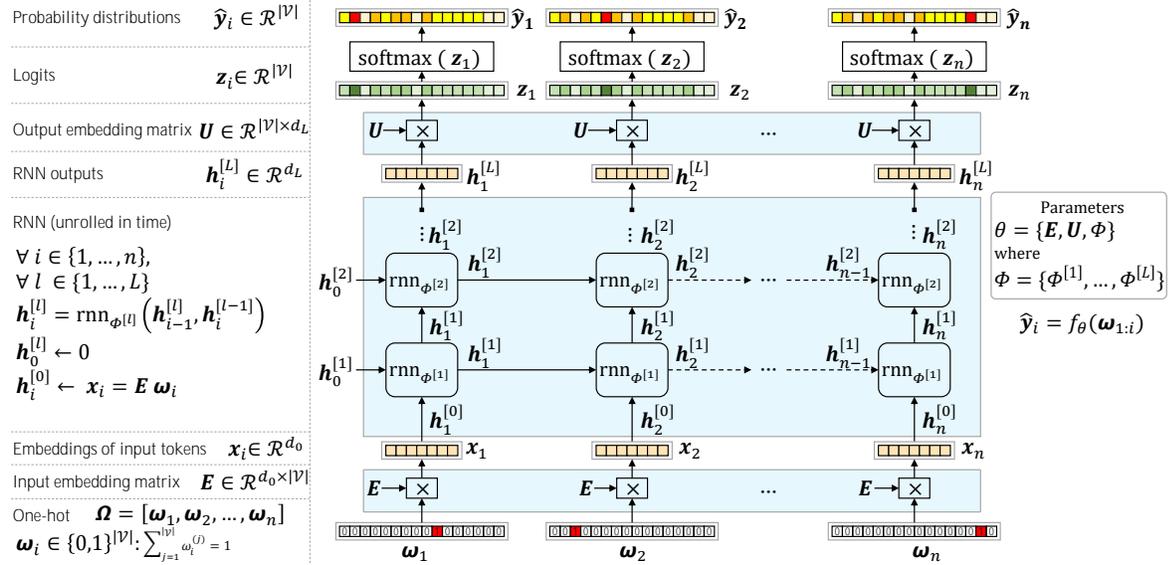

Figure 7. RNN language model

As with FFNN and transformer LMs, RNN outputs $h_i^{[L]}$ are then multiplied by the trainable output embedding matrix $U \in \mathcal{R}^{|\mathcal{V}| \times d_L}$ in order to obtain the logits $z_i \in \mathcal{R}^{|\mathcal{V}|}$ as shown in equation 9. Finally, the probability distributions $\hat{y}_i \in \mathcal{R}^{|\mathcal{V}|}$ are produced from the logits using the softmax function as shown in equation 10.

Encapsulating the parameters of the RNN LM as $\theta = \{E, U, \Phi\}$ with $\Phi = \{\Phi^{[1]}, \dots, \Phi^{[L]}\}$, the probability distributions $\hat{y}_i$ at each time step can be expressed as:

$$\hat{y}_i = f_\theta(\omega_{1:i}) \qquad (13)$$

The last function will be discussed in detail for two main architectures of RNNs i.e. Elman net [16] (the simple classical RNN) and the Long Short Term Memory (LSTM) [23] networks in Section 5.3.

### 4.5 Tying input and output embedding matrices

In practice, RNN (including LSTM) and transformer blocks in language models generate vectors of the same dimensionality as their input $d_l = d_e \; \forall\, l \in \{0, \dots, L\}$. Given this, the dimensionality of the input and output embedding matrices are given as $E \in \mathcal{R}^{d_e \times |\mathcal{V}|}$ and $U \in \mathcal{R}^{|\mathcal{V}| \times d_e}$, respectively. In fact, as observed in [46], $E$ and $U$ play a similar role. Both should capture semantic similarity between words. The columns of $E$ that represent similar words should be similar. Equivalently, the rows of $U$ that represent similar words should be similar. To this end, in the context of LSTM LMs, Press et al. [46] recommended tying the input embeddings with the output ones. In other words, they proposed to use one embedding matrix $E$ and to enforce $U = E^\mathsf{T}$. This results in two main advantages: 1) a considerable reduction in the number of trainable parameters which leads to reducing the computational cost and facilities the training, and 2) improving language modelling performance.

Weight tying was adopted in the original transformer work [58] and it is used in all transformer-based models. Therefore, we will make the following assumptions in the sequel for RNN and transformer LMs (but not the FFNN LM):

1. The input and output dimensionality of all the NN layers are the same. Let $d_e$ denote this dimension, then $d_l = d_e \ \forall \ l \in \{0, ..., L\}$. In fact, this is unavoidable in transformer LMs since they use residual connections as will be shown in Section 5.2.
2. Input and output embedding matrices are tied: $U = E^\mathsf{T}$.

### 4.6 Training autoregressive language models

As explained in the previous sections, given a context of $n$ tokens, transformer- and RNN- based language models generate a probability distribution over the vocabulary $\hat{y}_i^{(j)} = p_\theta(v_j|w_{1:i}) \ \forall \ v_j \in \mathcal{V}$ for all the indices $i \in \{1, ..., n\}$ of input tokens. This latter gives the estimated conditional probability of having the word $v_j$ as a next word, $w_{i+1}$, given the left side context $w_{1:i}$. Thus, at their output, $n$ probability distributions are generated with each of them conditioned on the current context $w_{1:i}$. This has been illustrated in Figure 5 and Figure 7. On the contrary, FFNNs generate only one probability distribution that expresses $\hat{y}_n^{(j)} = p_\theta(v_j|w_{1:n}) \ \forall \ v_j \in \mathcal{V}$. In the sequel, the general case of the probability distribution, $\hat{y}_i^{(j)}$, will be considered. This covers the transformer and RNN cases while simply replacing $\hat{y}_i^{(j)}$ by $\hat{y}_n^{(j)}$ makes the same explanation valid for FFNNs.

Language models are trained in a self-supervised way. When we train the model to predict the next word given a context $w_{1:i}$, the word $w_{i+1}$ that follows $w_{1:i}$ in the considered training sequence (and uniquely it) will be used as a ground truth. Clearly, no need for manually labeled data in this self-supervised learning framework and this enables using large language corpora as training sets.

The probability distribution estimated by a neural language model is represented by a $|\mathcal{V}|$-dimensional real-valued vector $\hat{y}_i$ whose components are given in equation 14:

$$\hat{y}_i = \left[\hat{y}_i^{(1)}, \hat{y}_i^{(2)}, ..., \hat{y}_i^{(|\mathcal{V}|)}\right]^\mathsf{T} \quad \forall \ i \in \{1, ..., n\}$$
$$\hat{y}_i^{(j)} = p_\theta(v_j|w_{1:i}) \quad \forall \ v_j \in \mathcal{V} \tag{14}$$

On the other hand, the true probability distribution is represented by a $|\mathcal{V}|$-dimensional one-hot vector $y_i \in \{0,1\}^{|\mathcal{V}|} : \sum_{j=1}^{|\mathcal{V}|} y_i^{(j)} = 1$ as given by equation 15:

$$y_i^{(j)} = p(v_j|w_{1:i}) = \begin{cases} 1, & v_j = w_{i+1} \\ 0, & \text{otherwise} \end{cases}, \forall \ v_j \in \mathcal{V} \tag{15}$$

Training autoregressive language models is based on minimizing the difference between the estimated probability distribution $p_\theta$ and the true one $p$. This is achieved by minimizing the cross entropy between the two probability distributions as depicted in Figure 8.

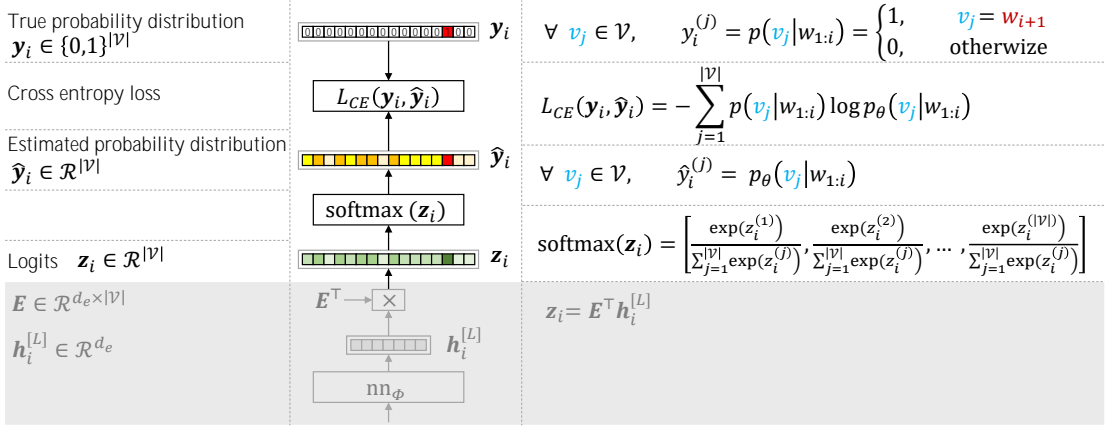

Figure 8. Cross entropy loss comparing the true probability distribution $p$ with the estimated one $p_\theta$

The cross entropy between the two discrete probability distributions $p_\theta$ and $p$ is given by definition in equation 16:
$$L_{CE}(\mathbf{y}_i, \hat{\mathbf{y}}_i) = -\sum_{j=1}^{|\mathcal{V}|} p(v_j|w_{1:i}) \log p_\theta(v_j|w_{1:i}) = -\sum_{j=1}^{|\mathcal{V}|} y_i^{(j)} \log \hat{y}_i^{(j)} \qquad (16)$$

Since $\mathbf{y}_i$ is a one-hot vector, equation 16 is reduced to:
$$L_{CE}(\mathbf{y}_i, \hat{\mathbf{y}}_i) = -\log p_\theta(v_{c_i}|w_{1:i}) = -\log y_i^{(c_i)} \qquad (17)$$

where $c_i \in \{1, \dots, |\mathcal{V}|\}$ represents the index of the correct token $v_{c_i} = w_{i+1}$ (among other tokens in the vocabulary). Thus, the components of the one-hot vector $\mathbf{y}_i$ can be represented in terms of $c_i$ as follows:
$$y_i^{(j)} = \begin{cases} 1, & j = c_i \\ 0, & \text{otherwise} \end{cases}, \forall j \in \{1, \dots, |\mathcal{V}|\} \qquad (18)$$

FFNN, RNN and transformer language models are trained using the cross entropy loss. However, they employ it in different ways inherited from each network architecture. In FFNN LMs, training is based on the iterative minimization of the following cost function with respect to the trainable parameters $\theta$:
$$\mathcal{J}_{AR}(\mathbf{y}_n, \hat{\mathbf{y}}_n) = L_{CE}(\mathbf{y}_n, \hat{\mathbf{y}}_n) = -\log y_n^{(c_n)} \qquad (19)$$

where $\hat{\mathbf{y}}_n = f_\theta(\mathbf{\Omega})$, recall that $\mathbf{\Omega}$ represents the one-hot vector representation of the input tokens.

Transformer LMs generate at the same time an output sequence of the same length, $n$, as of the input sequence. Therefore, they are trained to minimize the following loss function:
$$\mathcal{J}_{AR}(\mathbf{Y}, \hat{\mathbf{Y}}) = \sum_{i=1}^{n} L_{CE}(\hat{\mathbf{y}}_i, \mathbf{y}_i) = -\sum_{i=1}^{n} \log \hat{y}_i^{(c_i)} \qquad (20)$$

where $\hat{\mathbf{Y}} = [\hat{\mathbf{y}}_1, \hat{\mathbf{y}}_2, \dots, \hat{\mathbf{y}}_n] = f_\theta(\mathbf{\Omega})$.

In RNN LMs, $L_{CE}(\mathbf{y}_i, \hat{\mathbf{y}}_i)$ is minimized at each time step, i.e. the following loss function is iteratively minimized:
$$\mathcal{J}_{AR}(\mathbf{y}_i, \hat{\mathbf{y}}_i) = L_{CE}(\mathbf{y}_i, \hat{\mathbf{y}}_i) = -\log y_i^{(c_i)} \qquad (21)$$

where $\hat{\mathbf{y}}_i = f_\theta(\boldsymbol{\omega}_{1:i})$.

Note that for RNN and transformer LMs, the interim predicted tokens are not taken into account for predicting the next tokens. We always use the original sub-sequence $w_{1:i}$ as a context to predict the next token $w_{i+1}$. This technique is called teacher forcing.

Concerning the optimization technique, the gradient descent algorithm and its variants like ADAM [27] are widely used to minimize the loss function: $\underset{\theta}{\text{minimize}}\, \mathcal{J}_{AR}$. The gradient, $\nabla_\theta \mathcal{J}_{AR}$, of the cross entropy-based loss $\mathcal{J}_{AR}$ with respect to all the trainable parameters of the language model $\theta$ is computed.

In the gradient descent algorithm, the parameters are updated using the following rule:
$$\theta \leftarrow \theta - \mu_{lr} \nabla_\theta \mathcal{J}_{AR} \qquad (22)$$

where $\mu_{lr}$ represents the learning rate (step size). The gradient $\nabla_\theta \mathcal{J}_{AR}$ is computed using the chain rule. Deep learning frameworks like TensorFlow and PyTorch have highly optimized and accurate implementations of automatic differentiation. For a detailed description of the gradient descent algorithm, the reader is referred to [42 chapters 2 and 3] and [8 chapter 2]. For practical information on automatic differentiation, the reader is referred to [52 chapter 5].

## 4.7 The limits of input sequence length

The sequence length that can be processed by language models has several restrictions that come from the model architecture and/or the training algorithm. FFNN LMs are designed to receive fixed length sequences. Given that the model architecture is based on concatenating all the input embeddings and that the employed layers are dense, the feasible sequence length (with which model can be trained) is quite limited. Transformer LMs are also designed to receive fixed length sequences. However, this fixed length can be quite long and sufficient for most of NLP applications. For instance, the supported context length in GPT4 [43] can reach 32K tokens which is equivalent to 50 pages approximately. This impressive capacity comes from two main factors. The first is the ability of the attention mechanism to capture long term relationships between input tokens and thus to exploit the context efficiently. The second is that operations in transformers are parallelized which enable a fast and efficient training. The attention mechanism and the parallel architecture of transformers will be explored in detail in Section 5.2. RNN LMs theoretically support unlimited sequence length. However, in practice, the feasible length is restricted by the following factors: 1) the gradient of the loss function with respect to the trainable parameters engages all the previous steps of the recurrent network. This means that the sequence length is limited by the capacity of the automatic differentiation computational graph. 2) The sequential way of processing the sequence makes training inefficient compared to transformer LMs. 3) the recurrence mechanism suffers from the vanishing/exploding gradient problem that limits the ability of capturing long term dependencies and thus limits the supported sequence length. This problem will be discussed in detail in Section 5.3.

## 4.8 The order of input tokens

In FFNN LMs, the order of input tokens is implicitly taken into account by the concatenation step of the input embeddings. The resultant features' vector is not orderless since the position of its components will correspond to particular weights (neurons) in the fully connected layers. In transformer LMs, the order is not natively taken into account since the operations of the transformer blocks are orderless i.e. if we shuffle the input vectors of a transformer block, the corresponding outputs will not change. To cope with this issue, positional encodings are explicitly added to the input embeddings. Thanks to the residual connections, these positional information can travel across multiple stacked blocks as will be explored in Section 5.2. RNN LMs support natively the order of input tokens because they process the sequence sequentially token by token as will be shown in Section 5.3.

## 5 NEURAL NETWORKS IN AUTOREGRESSIVE LANGUAGE MODELS

After having explored the high-level framework of AR LMs, we provide in the following sections the detailed architecture of their neural components. Particularly, we explain how the outputs $\boldsymbol{h}_i^{[L]} \in \mathcal{R}^{d_e}$ in transformers and RNNs and $\boldsymbol{h}_n^{[L]} \in \mathcal{R}^{d_L}$ in FFNNs are generated from the embeddings of the input tokens $\boldsymbol{x}_i \in \mathcal{R}^{d_e} \ \forall i \in \{1, \dots, n\}$.

## 5.1 Feedforward neural networks in autoregressive language models

Figure 9 shows how a feedforward neural network $\text{ffnn}_\Phi$ generates a representation of the predicted token $\boldsymbol{h}_n^{[L]}$ starting from the embeddings of the input tokens $\boldsymbol{x}_i \in \mathcal{R}^{d_e} \ \forall \ i \in \{1, \dots, n\}$.

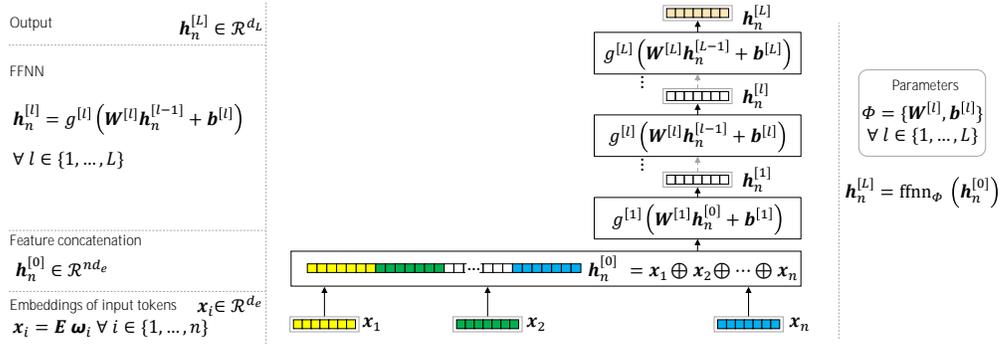

Figure 9. Feedforward neural network in AR LM

The first step is to concatenate the input embeddings in one vector $h_n^{[0]} \in \mathcal{R}^{nd_e}$. Then, $L$ fully connected layers are applied as given by equation 23:

$$h_n^{[l]} = g^{[l]}\left(W^{[l]} h_n^{[l-1]} + b^{[l]}\right) \quad \forall\, l \in \{1, \dots, L\} \tag{23}$$

where the weight matrix $W^{[l]}$ and the bias vector $b^{[l]}$ are trainable parameters and $g^{[l]}$ represents the activation function of the $l^{\text{th}}$ layer. The set of trainable parameters of $\text{ffnn}_\Phi$ is $\Phi = \{W^{[l]}, b^{[l]}\}\ \forall\, l \in \{1, \dots, L\}$. The original FFNN LM of Bengio [2] contained one hidden layer and used the sigmoid activation function $g^{[1]}(x) = \sigma(x) = 1/(1 + e^{-x})$.

## 5.2 Transformer Networks in autoregressive language models

In this section, we explain how a transformer transforms a sequence $H^{[0]} = [h_1^{[0]}, h_2^{[0]}, \dots, h_n^{[0]}] \in \mathcal{R}^{d_e \times n}$ to a new sequence of the same length $H^{[L]} = [h_1^{[L]}, h_2^{[L]}, \dots, h_n^{[L]}] \in \mathcal{R}^{d_e \times n}$. Particularly, the function $H^{[L]} = \text{transformer}_\Phi(H^{[0]})$ given in equation 8 will be described in detail. Note that transformers can be configured to be used in AR LMs where only left side context is accessible or in bidirectional LMs where the context from both left and right sides can be accessed. Depending on the context-access configuration, a transformer block is called either a "transformer decoder" or a "transformer encoder". In fact, in AR configurations, a transformer block is called a transformer decoder because it can be used for text generation while a bidirectional transformer block is called a transformer encoder[5] [12]. Therefore, we will describe in this section the generic transformer block and we will show how it can be configured to satisfy the auto-regression or autoencoding conditions. We will start by exploring the basic building blocks of the transformer. Then we build them up step by step to constitute $\text{transformer}_\Phi$.

The main idea underlying transformers is to systematically analyze the relationships between different tokens in a given sequence. This is achieved by what's called the self-attention mechanism [58]. This mechanism was formulated using the notion of the dictionary (or associative array) data structure used in programming languages and databases [55 chapter 5]. A dictionary is a set of key-value pairs. It is indexed by unique keys. One of the main operations on a dictionary is to extract a value that correspond to a given key. Thus, given a query, a dictionary returns the value corresponding to the key that exactly matches the query. In the self-attention mechanism, the concept of key, value and query is extended where these latter are represented by real-valued vectors. Instead of matching a query against a key, a similarity score between these two vectors, that should have the same dimensionality, is calculated using vector dot product. Based on this concept, let us explore how the attention mechanism works through an example.

---

[5] Some transformer networks use both encoders and decoders like the ones used in automatic translation. They belong to the family of encoder-decoder networks (or equivalently sequence-to-sequence networks). This family is out of the scope of this tutorial which focuses on language modeling.

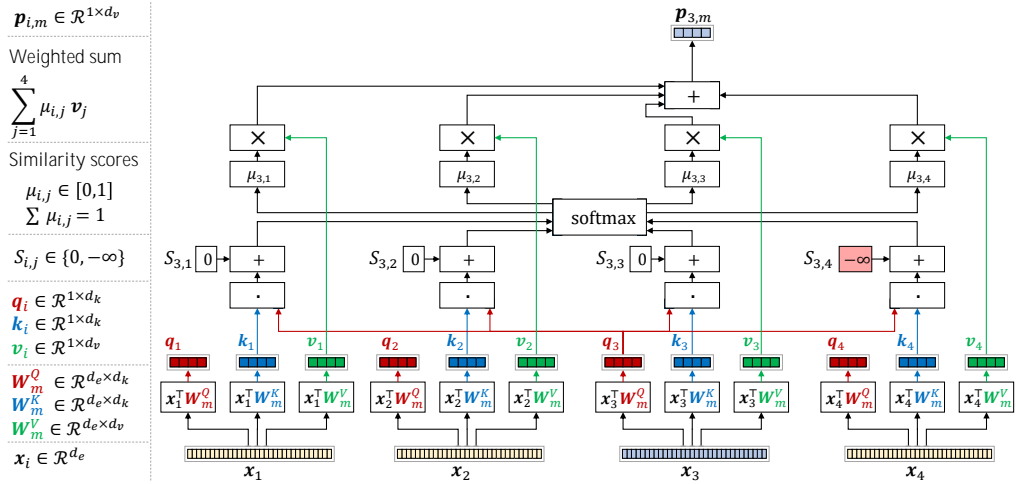

Figure 10. Self-attention explained by example

Assume we have a sequence of 4 tokens represented by dense embeddings $X = [x_1, x_2, x_3, x_4]$ where $x_i \in \mathcal{R}^{d_e} \; \forall i \in \{1,2,3,4\}$. Recall that the objective of AR language modeling is to predict the next word given its left-side context. Let us consider for example the problem of predicting the 4th token given the first 3 ones. Figure 10 shows how self-attention manipulates the input tokens in order to generate an intermediate representation $p_{3,m}$ that serves the AR language modeling objective. Note that the focus in this example is set on the 3rd token: the place where we want to predict the next token. Starting from each embedding vector $x_i$, self-attention generates three vectors $q_i \in \mathcal{R}^{1 \times d_k}$, $k_i \in \mathcal{R}^{1 \times d_k}$, and $v_i \in \mathcal{R}^{1 \times d_v}$ representing the query, key and value vectors, respectively. Query and key vectors have the same dimensions $d_k$ while the dimension of the value vector $d_v$ can be different. These vectors are given as follows:

$$\begin{aligned} q_i &= x_i^\top W_m^Q \\ k_i &= x_i^\top W_m^K \\ v_i &= x_i^\top W_m^V \end{aligned} \qquad (24)$$

where $W_m^Q \in \mathcal{R}^{d_e \times d_k}$, $W_m^K \in \mathcal{R}^{d_e \times d_k}$ and $W_m^V \in \mathcal{R}^{d_e \times d_v}$ represent three trainable matrices.

With the focus set on $x_3$, the considered query is $q_3$ which will be compared against all the produced keys. The similarity between the query $q_3$ and each of the keys $k_i$ is calculated by means of vector dot product. Since the problem at hand is autoregressive language modeling, $x_4$ should not be accessed. This is managed by a mask $S_{i,j} \in \{0, -\infty\}$ that removes the contribution of the right context tokens from the intermediate representation $p_{3,m}$. In this example, $S_{3,4} = -\infty$ and $S_{3,i} = 0 \; \forall i \in \{1,2,3\}$. After adding this mask to the resultant similarity scores, the softmax function is used to produce normalized scores $\mu_{i,j} \in [0,1], \sum \mu_{i,j} = 1$ as given in equation 25:

$$\mu_3 = \text{softmax}([S_{3,1} + q_3 \cdot k_1, S_{3,2} + q_3 \cdot k_2, S_{3,3} + q_3 \cdot k_3, S_{3,4} + q_3 \cdot k_4]) = \\ \left[ \frac{\exp(S_{3,1}+q_3 \cdot k_1)}{\sum_{j=1}^{4} \exp(S_{3,j}+q_3 \cdot k_j)}, \frac{\exp(S_{3,2}+q_3 \cdot k_2)}{\sum_{j=1}^{4} \exp(S_{3,j}+q_3 \cdot k_j)}, \frac{\exp(S_{3,3}+q_3 \cdot k_3)}{\sum_{j=1}^{4} \exp(S_{3,j}+q_3 \cdot k_j)}, \frac{\exp(S_{3,4}+q_3 \cdot k_4)}{\sum_{j=1}^{4} \exp(S_{3,j}+q_3 \cdot k_j)} \right] \qquad (25)$$

Note that when $S_{i,j} = -\infty$, the corresponding score will be zeroed out. This handy mask enables configuring the transformer block to work in AR LM and in masked language models that will be explored in Section 6.1.

The scores are then multiplied by the value vectors to generate the self-attention result:

$$p_{i,m} = \sum_{j=1}^{4} \mu_{i,j} v_j \qquad (26)$$

Transformers use different randomly initialized versions of the trainable matrices $W_m^Q, W_m^K, W_m^V \; \forall m \in \{1, \ldots, M\}$ in order to capture different types of relationships between the input embeddings. The self-attention mechanism with one set of these matrices (i.e. for a given $m$) is called an attention-head. The operation employing all the attention heads is called multi-head attention. This latter is depicted in Figure 11.

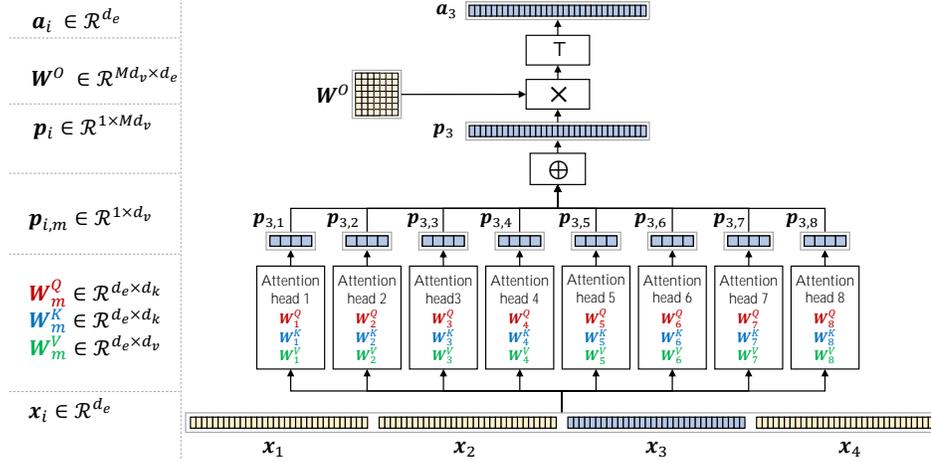

Figure 11. Multi-head attention explained by example

The results of different attention heads $\boldsymbol{p}_{i,m} \in \mathcal{R}^{1 \times d_v}$ are horizontally concatenated in one vector $\boldsymbol{p}_i \in \mathcal{R}^{1 \times Md_v}$ where $M$ represents the number of attention heads. To project back $\boldsymbol{p}_i \in \mathcal{R}^{1 \times Md_v}$ to the vector space $\mathcal{R}^{d_e}$, the following equation is used:

$$\boldsymbol{a}_3 = (\boldsymbol{p}_3 \boldsymbol{W}^O)^\top \qquad (27)$$

where $\boldsymbol{W}^O \in \mathcal{R}^{Md_v \times d_e}$ represents a trainable parameter. In practice, an attention head is applied on the entire sequence using matrix operations. Let us consider the general case where the self-attention mechanism is applied to a sequence $\boldsymbol{X} = [\boldsymbol{x}_1, \boldsymbol{x}_2, \dots, \boldsymbol{x}_n]$ of $n$ vectors (either the input embeddings or any intermediate "contextualized" representation of the tokens). The computation flow of self-attention is depicted in Figure 12. Self-attention, for the $m^{\text{th}}$ attention head, is expressed as a parametrized function $\boldsymbol{P}_m = \text{selfAttention}_{\boldsymbol{\Psi}_m}(\boldsymbol{X})$ whose parameters are $\boldsymbol{\Psi}_m = \{\boldsymbol{W}_m^Q, \boldsymbol{W}_m^K, \boldsymbol{W}_m^V\}$. This function maps the input sequence $\boldsymbol{X} \in \mathcal{R}^{d_e \times n}$ to a matrix $\boldsymbol{P}_m \in \mathcal{R}^{n \times d_v}$ as given by equation 28[6]:

$$\boldsymbol{P}_m = \text{selfAttention}_{\boldsymbol{\Psi}_m}(\boldsymbol{X}) = \text{softmax}\left(\boldsymbol{S} + \frac{\boldsymbol{X}^\top \boldsymbol{W}_m^Q \boldsymbol{W}_m^{K\top} \boldsymbol{X}}{\sqrt{d_k}}\right) \boldsymbol{X}^\top \boldsymbol{W}_m^V \qquad (28)$$

where the mask $\boldsymbol{S} \in \mathcal{R}^{n \times n}$ is given by equation 29:

$$\boldsymbol{S} = \begin{cases} \begin{bmatrix} 0 & -\infty & -\infty & & -\infty & -\infty & -\infty \\ 0 & 0 & -\infty & \dots & -\infty & -\infty & -\infty \\ 0 & 0 & 0 & & -\infty & -\infty & -\infty \\ & \vdots & & & \vdots & & \\ 0 & 0 & 0 & & 0 & -\infty & -\infty \\ 0 & 0 & 0 & \dots & 0 & 0 & -\infty \\ 0 & 0 & 0 & & 0 & 0 & 0 \end{bmatrix} & \text{for AR LM} \\ \begin{bmatrix} 0 & 0 & 0 & & 0 & 0 & 0 \\ \vdots & \vdots & \vdots & \dots & \vdots & \vdots & \vdots \\ 0 & 0 & 0 & & 0 & 0 & 0 \end{bmatrix} & \text{for AE LM} \end{cases} \qquad (29)$$

---

[6] This equation is equivalent to the equation $\text{selfAttention}(\boldsymbol{Q}, \boldsymbol{K}, \boldsymbol{V}) = \text{softmax}\left(\frac{\boldsymbol{Q}\boldsymbol{K}^\top}{\sqrt{d_k}}\right)\boldsymbol{V}$ that is given in [58]. However, for the sake of clarity, our notation makes an explicit difference between the trainable parameters of a function and its input arguments. In addition, we integrated the mask $\boldsymbol{S}$ in the equation to generalize for AR and AE LMs.

Note that the scaling factor $\sqrt{d_k}$ in equation 28 is used to improve the numerical conditioning of the query-key dot products [58].

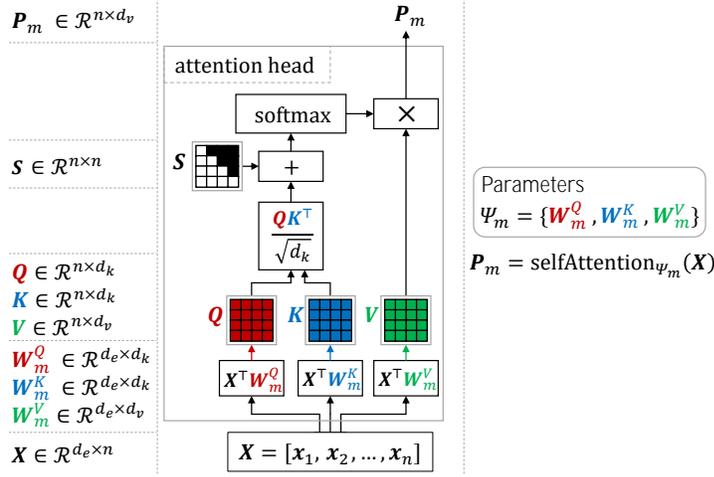

Figure 12. Self-attention head

Based on the definition of $\text{selfAttention}_{\Psi_m}$, Figure 13 shows how multi-head attention works. Simply, it applies $\text{selfAttention}_{\Psi_m} \forall m \in \{1, ..., M\}$ on its input sequence to generate the matrices $P_m = \mathcal{R}^{n \times d_v} \forall m \in \{1, ..., M\}$. These latter are horizontally concatenated in one matrix $P \in \mathcal{R}^{n \times M d_v}$. Then, $P$ is projected back to the space $\mathcal{R}^{d_e \times n}$ as follows:

$$A = (P W^O)^\top \qquad (30)$$

where $W^O \in \mathcal{R}^{M d_v \times d_e}$ is a trainable parameter. The multi-head attention function is represented in equation 31:

$$A = \text{multiHeadAttention}_\Gamma(X) \qquad (31)$$

where $\Gamma = \{W^O, \Psi_m\}, \forall m \in \{1, ..., M\}$ represents its trainable parameters' set, recall that $\Psi_m = \{W_m^Q, W_m^K, W_m^V\}$.

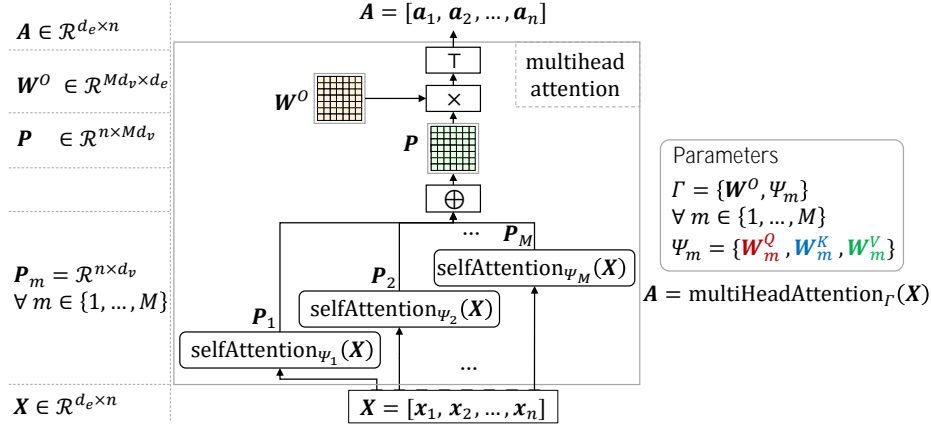

Figure 13. Multi-head attention

Let us explore now how multi-head attention is used, besides other components, in a complete transformer block. A transformer block is illustrated in Figure 14-a. It takes as input the intermediate representations from a previous block $H^{[l-1]} \in \mathcal{R}^{d_e \times n}$ (or the input representations $H^{[0]} = X + \Lambda \in \mathcal{R}^{d_e \times n}$) and returns a sequence of the same length $H^{[l]} \in \mathcal{R}^{d_e \times n}$. Thus, it is represented by the following function:

$$H^{[l]} = \text{transBlock}_{\Phi^{[l]}}(H^{[l-1]}) \tag{32}$$

where $\Phi^{[l]}$ represents its trainable parameters that will be detailed in the sequel. A transformer block starts by applying the multi-head attention function on its input as given in equation 33 (the same as equation 31 but here we recovered the layer index that have been dropped in the previous steps for the sake of simplicity):

$$A^{[l]} = \text{multiHeadAttention}_{\Gamma^{[l]}}(H^{[l-1]}) \tag{33}$$

Then, a residual connection operation is applied i.e. the output of **multiHeadAttention** is cumulated with its input. Residual connections are based on the hypothesis that the residual mapping is easier to optimize compared to the original unreferenced mapping [21]. In practice, they facilitate the training of deeper neural networks [21]. The result $H^{[l-1]} + A^{[l]}$ is then passed through a layer normalization function [1]:

$$C^{[l]} = \text{layerNorm}_{\Xi_1^{[l]}}(H^{[l-1]} + A^{[l]}) \tag{34}$$

where $\Xi_1^{[l]} = \{\alpha_1^{[l]}, \beta_1^{[l]}\}$ with $\alpha_1^{[l]} \in \mathcal{R}^{d_e}$; $\beta_1^{[l]} \in \mathcal{R}^{d_e}$ represents its set of trainable parameters. Note that **layerNorm** in equation 34 is applied column-wise to its input matrix. For a given column vector $\varkappa \in \mathcal{R}^{d_e}$, **layerNorm** is given in equation 35:

$$\text{layerNorm}_{\{\alpha,\beta\}}(\varkappa) = \alpha \odot \varkappa' + \beta \tag{35}$$

where $\varkappa' = \frac{\varkappa - \mu}{\sigma}$ is the z-scored version of $\varkappa$ (zero mean, unit variance vector), with $\mu = \frac{1}{d_e}\sum_{i=1}^{d_e} \varkappa_i$ and $\sigma = \sqrt{\frac{1}{d_e}\sum_{i=1}^{d}(\varkappa_i - \mu)^2}$, while the gain $\alpha \in \mathcal{R}^{d_e}$ and the bias $\beta \in \mathcal{R}^{d_e}$ are two trainable parameters.

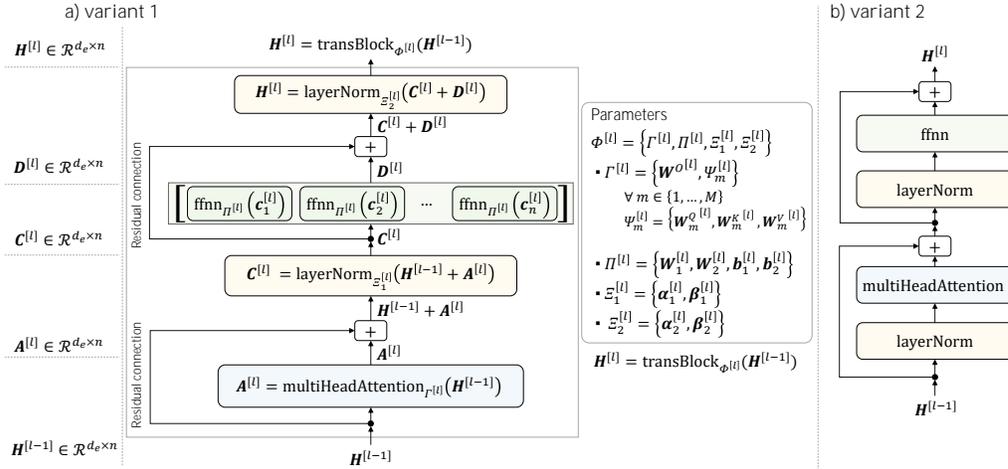

Figure 14. Transformer block: a) variant 1 and b) variant 2

The next step is to apply a feedforward network to the columns of $C^{[l]}$ i.e. $c_i^{[l]}\ \forall\ i \in \{1,...,n\}$ as given in equation 36:

$$D^{[l]} = \left[\text{ffnn}_{\Pi^{[l]}}(c_1^{[l]}), \text{ffnn}_{\Pi^{[l]}}(c_2^{[l]}), ..., \text{ffnn}_{\Pi^{[l]}}(c_n^{[l]})\right] \tag{36}$$

where $\text{ffnn}_{\Pi^{[l]}}$ is given by:

$$d_i^{[l]} = \text{ffnn}_{\Pi^{[l]}}(c_i^{[l]}) = W_2^{[l]}\text{GELU}\left(W_1^{[l]}c_i^{[l]} + b_1^{[l]}\right) + b_2^{[l]} \quad \forall i \in \{1,2,...,n\} \tag{37}$$

with $\Pi^{[l]} = \{W_1^{[l]}, W_2^{[l]}, b_1^{[l]}, b_2^{[l]}\}$ representing the set of trainable parameters of the FFNN network and **GELU** represents the Gaussian Error Linear Unit activation function [22]. This latter is given by $\text{GELU}(x) = xP(X \leq x)$ with $X \sim \mathcal{N}(0,1)$ and is usually approximated by the efficient formula $\text{GELU}(x) = 0.5\ x\ (1 + \tanh(\sqrt{2/\pi}(x + 0.044715\ x^3)))$ [22]. Note that the original work of Vaswani et al. [58] used the Rectified Linear Unit activation $\text{ReLU}(x) = \max(0, x)$. However, since later transformer-based models like BERT [12], GPT [4, 47, 48], ViT [14] and PatchTST [41] replaced ReLU by GELU, we adopted the latter in this tutorial. The interest of using the GELU activation function is to weight inputs by their value, rather than gating them by their sign as in ReLU [22].

The function $\text{ffnn}_{\Pi^{[l]}}$ (with the same parameters) is applied in parallel to all $c_i^{[l]}$ $\forall i \in \{1, \dots, n\}$. The dimensions of the trainable parameters are $W_1^{[l]} \in \mathcal{R}^{d_f \times d_e}$, $b_1^{[l]} \in \mathcal{R}^{d_f}$, $W_2^{[l]} \in \mathcal{R}^{d_e \times d_f}$ and $b_2^{[l]} \in \mathcal{R}^{d_e}$ where $d_f$ is called the feedforward dimension or the internal dimension to differentiate it from the embedding dimension $d_e$ which is called the model dimension or the external dimension in some references.

A residual connection is applied to the output of the FFNN. Finally, the result $C^{[l]} + D^{[l]}$ is fed to a layer normalization step as given in equation 38:

$$H^{[l]} = \text{layerNorm}_{\Xi_2^{[l]}}(C^{[l]} + D^{[l]}) \tag{38}$$

where $\Xi_2^{[l]} = \{\alpha_2^{[l]}, \beta_2^{[l]}\}$ represents the corresponding set of trainable parameters. The complete set of trainable parameters $\Phi^{[l]}$ is shown at the right side of Figure 14.

In practice, a transformer $H^{[L]} = \text{transformer}_\Phi(H^{[0]})$ consists of a stack of transformer blocks $H^{[l]} = \text{transBlock}_{\Phi^{[l]}}(H^{[l-1]})$ $\forall l \in \{1, \dots, L\}$ as depicted in Figure 15. The set of its trainable parameters is $\Phi = \{\Phi^{[1]}, \Phi^{[2]}, \dots \Phi^{[L]}\}$.

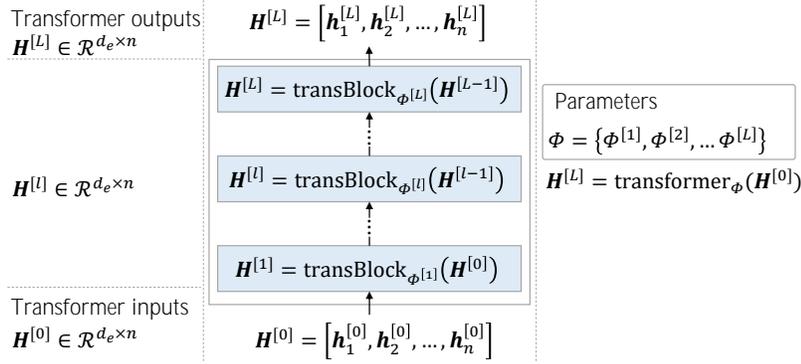

Figure 15. Deep transformer with $L$ stacked blocks

A transformer-based AR language model $\hat{Y} = f_\theta(\Omega)$ is illustrated in Figure 16 where $\theta = \{E, \Lambda, \Xi_e, \Phi\}$ represents its trainable parameters' set. The weight tying is applied where the same embedding matrix $E$ is used to represent input and output embeddings. Intermediate representations $H^{[l]}$ $\forall l \in \{1, \dots, L\}$ are normalized using the **layerNorm** functions in transformer blocks. In the same way, the input $(\Lambda + X)$ is normalized using embedding layer normalization given in equation 39:

$$H^{[0]} = \text{layerNorm}_{\Xi_e}(\Lambda + X) \tag{39}$$

where $\Xi_e = \{\alpha_e, \beta_e\}$ represents the corresponding set of trainable parameters. Note that in a transformer block shown in Figure 14, the layer normalization is applied after the **multiHeadAttention**$_{\Gamma^{[l]}}$ and after the **ffnn**$_{\Pi^{[l]}}$. In some implementations like the implementation of GPT2 [48] in KerasNLP, layer normalization is applied before the aforementioned processing steps as shown in Figure 14-b. In such configurations, **layerNorm**$_{\Xi_e}$ must be placed at the output of the transformer instead of placing it at its input in Figure 16.

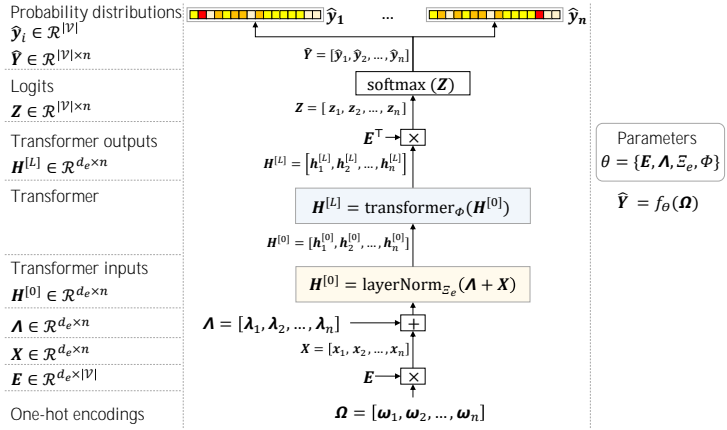

Figure 16. Transformer AR LM

Figure 17 shows, from up to bottom, all the trainable parameters of an AR LM. Parameters shown in brown $b^{O[l]} \in \mathcal{R}^{d_e}, b_m^{Q\,[l]} \in \mathcal{R}^{d_k}, b_m^{K\,[l]} \in \mathcal{R}^{d_k}$ and $b_m^{V\,[l]} \in \mathcal{R}^{d_v}$ are optional. They represent bias vectors added to the results of linear transformations made by the multi-head attention function. The hyper parameters are also presented in this figure which are the embedding dimension $d_e$, the value vector dimension $d_v$, the key vector dimension $d_k$, the feedforward dimension $d_f$, the number of attention heads $M$, the number of transformer blocks $L$, the vocabulary size $|\mathcal{V}|$ and the maximum sequence length $n$.

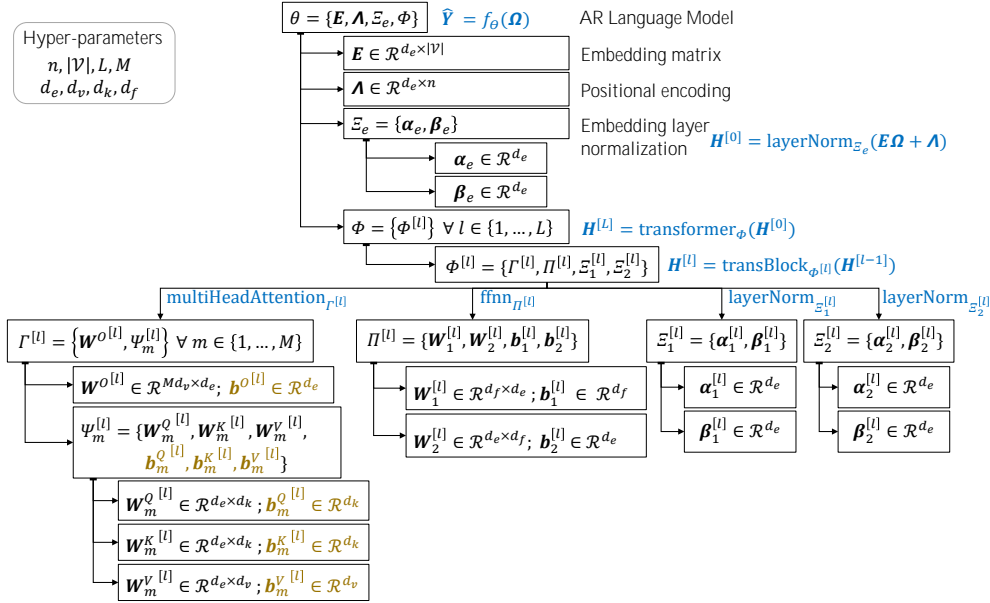

Figure 17. Parameters of transformer AR LM

Based on the set of trainable parameters and their corresponding dimensionalities given in Figure 17, the detailed calculation of the number of trainable parameters is derived in Table 4. The variable $\zeta \in \{0,1\}$ indicates whether the optional trainable parameters shown in Figure 17 are considered or not.

Based on Table 4, the total number of trainable parameters of the GPT2 LM is given by equation 40:

$$\eta_{\text{GPT2}} = d_e(|\mathcal{V}| + n + 2) + L\big(2Md_e(d_k + d_v) + \zeta(M(2d_k + d_v) + d_e) + 2d_e d_f + 5d_e + d_f\big) \tag{40}$$

We implemented these formulas and the ones in Section 6.1 in Python and they are publically available on the link indicated in the introduction. These formulas give exactly the same number of trainable parameters obtained by parameters' counters utilities of TensorFlow. We also implemented forward pass of GPT2 from scratch and validated that it gives identical results compared to the KerasNLP implementation.

Table 4, detailed calculation of the number of trainable parameters of GPT2 autoregressive language model.

| | |
|---|---|
| Embedding | $\eta_1 = d_e|\mathcal{V}|$ |
| Positional encoding | $\eta_2 = d_e n$ |
| layerNorm$_{\Xi_e}$ | $\eta_3 = 2d_e$ |
| multiHeadAttention$_{\Gamma^{[l]}}$ | $\eta_4 = 2Md_e(d_k + d_v) + \zeta(M(2d_k + d_v) + d_e)$ |
| ffnn$_{\Pi^{[l]}}$ | $\eta_5 = 2d_e d_f + d_e + d_f$ |
| layerNorm$_{\Xi_1^{[l]}}$ | $\eta_6 = 2d_e$ |
| layerNorm$_{\Xi_2^{[l]}}$ | $\eta_7 = 2d_e$ |
| transBlock$_{\Phi^{[l]}}$ | $\eta_8 = \eta_4 + \eta_5 + \eta_6 + \eta_7$ |
| transformer$_\Phi$ | $\eta_9 = L\eta_8$ |
| GPT2 AR LM | $\eta_{\text{GPT2}} = \eta_1 + \eta_2 + \eta_3 + \eta_9 = d_e(|\mathcal{V}| + n + 2) +$ $L\big(2Md_e(d_k + d_v) + \zeta(M(2d_k + d_v) + d_e) + 2d_e d_f + 5d_e + d_f\big)$ |

## 5.3 Recurrent neural networks in autoregressive language models

In this section, RNN-based LMs are explained. Two types of RNNs are considered: 1) simple RNN namely, Elman net [16], and 2) Long Short-Term Memory (LSTM) [23]. We will loosely refer to Elman nets as RNNs in the sequel.

Figure 18 shows an RNN unrolled in time. It consists of $L$ recurrent layers $\boldsymbol{h}_i^{[l]} = \text{rnn}_{\Phi^{[l]}}\big(\boldsymbol{h}_{i-1}^{[l]}, \boldsymbol{h}_i^{[l-1]}\big)$. At a given time step, $i$, the output of $\text{rnn}_{\Phi^{[l]}}$ depends on its input $\boldsymbol{h}_i^{[l-1]}$ which comes from the previous layer and on its pervious output $\boldsymbol{h}_{i-1}^{[l]}$. The inputs of the first layer are the embeddings $\boldsymbol{h}_i^{[0]} \leftarrow \boldsymbol{x}_i$ while the initial recurrent input is set to zero $\boldsymbol{h}_0^{[l]} \leftarrow 0$.

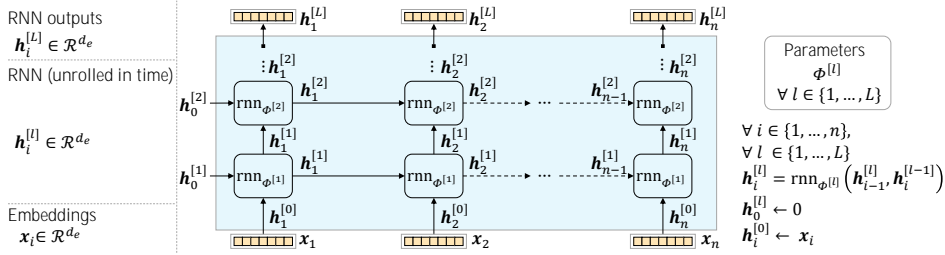

Figure 18. Recurrent neural network unrolled in time

Figure 19 shows an RNN layer. Its recurrent function $\text{rnn}_{\Phi^{[l]}}$ is given by equation 41:

$$\boldsymbol{h}_0^{[l]} \leftarrow \boldsymbol{0}; \quad \boldsymbol{h}_i^{[0]} \leftarrow \boldsymbol{x}_i$$
$$\boldsymbol{h}_i^{[l]} = \text{rnn}_{\Phi^{[l]}}\big(\boldsymbol{h}_i^{[l-1]}, \boldsymbol{h}_{i-1}^{[l]}\big) = g^{[l]}\big(\boldsymbol{U}^{[l]}\boldsymbol{h}_{i-1}^{[l]} + \boldsymbol{W}^{[l]}\boldsymbol{h}_i^{[l-1]} + \boldsymbol{b}^{[l]}\big) \quad \forall i > 0; \forall l \in \{1, \dots, L\} \tag{41}$$

where $\Phi^{[l]} = \{\boldsymbol{W}^{[l]}, \boldsymbol{U}^{[l]}, \boldsymbol{b}^{[l]}\}$ represents the set of its trainable parameters and $g^{[l]}$ is an activation function. In practice, $g^{[l]} = \tanh$ is commonly used in Elman RNNs. Note that the first RNN LM proposed by Mikolov et al. [38] consisted of one hidden layer with the sigmoid activation function and one output layer with the softmax function.

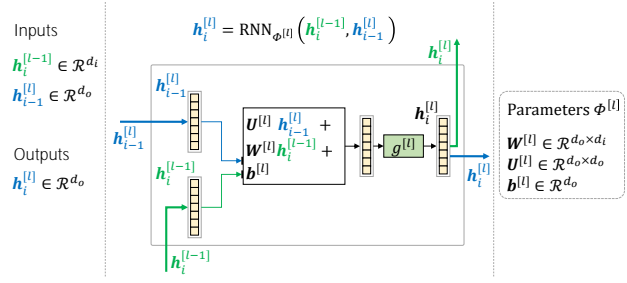

Figure 19. RNN layer

The dimensions of the trainable parameters are $W^{[l]} \in \mathcal{R}^{d_o \times d_i}$, $U^{[l]} \in \mathcal{R}^{d_o \times d_o}$, and $b^{[l]} \in \mathcal{R}^{d_o}$ with $d_i$ and $d_o$ representing the input and output dimensions, respectively. In practice, in order to facilitate stacking layers, we set $d_i = d_o = d_e$. Thus, the number of trainable parameters in one layer is given by equation 42:

$$\eta_{RNN} = 2d_e^2 + d_e \qquad (42)$$

RNN LMs have a major drawback which is the difficulty in learning long-term dependencies. Two factors underlying this problem are: 1) gradients might vanish or explode when the sequence length is large, and 2) RNN tries to manage, at the same time, two distinct problems: a) how to generate the best local predictions given the previous context, and b) how to store relevant information that will serve future decisions. While the first constraint is inherited from the recurrence mechanism, the second one is dependent on the RNN layer architecture. To overcome learning difficulties in RNNs, LSTMs were proposed. Figure 20 shows an LSTM network unrolled in time.

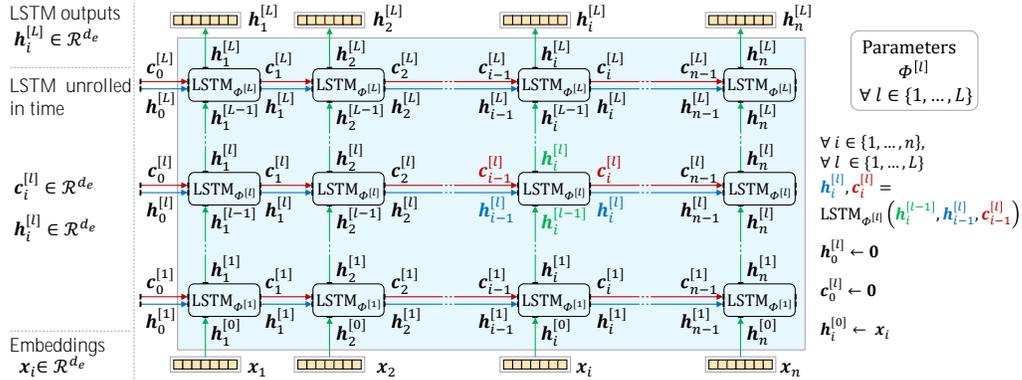

Figure 20. LSTM unrolled in time

Besides the classical recurrent signal $h_i^{[l]}$, LSTMs add an explicit context signal $c_i^{[l]}$. This separation of roles facilitates the management of the two main problems at hand: generating local decisions and memorizing information for future decisions. The figure shows that the outputs $\left(h_i^{[l]}, c_i^{[l]}\right)$ of an LSTM layer $\text{LSTM}_{\Phi^{[l]}}$ at a given time step, $i$, depend on the previous outputs $h_{i-1}^{[l]}$ and $c_{i-1}^{[l]}$ and on the current input $h_i^{[l-1]}$ that comes from the previous layer. The inputs of the first layer are the embeddings while the initial recurrent inputs are set to zero. This is expressed in equation 43:

$$\begin{aligned} h_i^{[l]}, c_i^{[l]} &= \text{LSTM}_{\Phi^{[l]}}\left(h_i^{[l-1]}, h_{i-1}^{[l]}, c_{i-1}^{[l]}\right) \\ h_0^{[l]} &\leftarrow \mathbf{0}; \qquad c_0^{[l]} \leftarrow \mathbf{0}; \qquad h_i^{[0]} \leftarrow x_i \end{aligned} \qquad (43)$$

Before diving in $\text{LSTM}_{\Phi^{[l]}}$, let us explore the main building block of this layer which is the multiplicative gate illustrated in Figure 21-a. Gating a vector $v \in \mathcal{R}^d$ can be understood as passing its relevant components (features) and suppressing the irrelevant ones. The relevance of components is determined by a semi-binary mask $y \in \mathcal{R}^d$ which results from applying the sigmoid function on a control vector $x \in \mathcal{R}^d$. Thus, the components of $y \in \mathcal{R}^d$ will be between zero and one and they will be dominated by values close to zero or to one, hence the name semi-binary mask. The gated vector $\tilde{v} \in \mathcal{R}^d$ is obtained by a component-wise multiplication of $v$ by the semi-binary mask i.e. $\tilde{v} = v \odot \sigma(x)$.

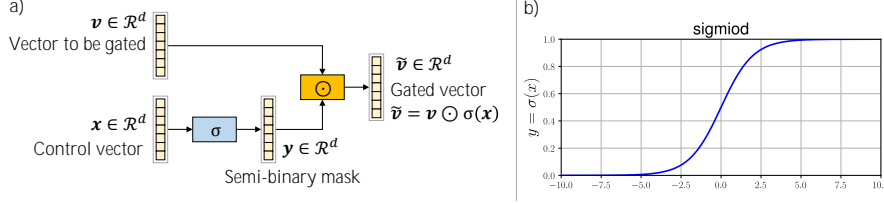

Figure 21. a) Multiplicative gate and b) the sigmoid function

Figure 22 illustrates the concept of multiplicative gating with an example on a randomly generated 20-dimensional vector. Figure 22-a shows the vector to be gated, Figure 22-b illustrates a random semi-binary mask and Figure 22-c shows the resultant gated vector.

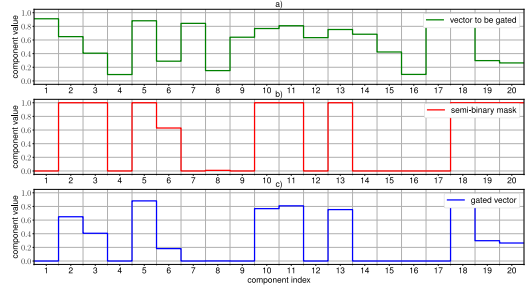

Figure 22. Illustration of the multiplicative gating concept: a) the vector to be gated, b) a semi-binary mask, and c) the resultant gated vector

This gating mechanism is used by LSTMs to manage what information to forget and what information to remember. Figure 23 shows an LSTM layer. The classical recurrent operation is given by equation 44:

$$q_i^{[l]} = \tanh\left(U^{Q[l]} h_{i-1}^{[l]} + W^{Q[l]} h_i^{[l-1]} + b^{Q[l]}\right) \tag{44}$$

Three gates are used to manage the context and the intermediate outputs. The first one, namely the forget gate, is given by equation 45:

$$\begin{aligned} p_i^{[l]} &= \sigma\left(U^{P[l]} h_{i-1}^{[l]} + W^{P[l]} h_i^{[l-1]} + b^{P[l]}\right) \\ k_i^{[l]} &= c_{i-1}^{[l]} \odot p_i^{[l]} \end{aligned} \tag{45}$$

The second gate, namely the add gate, is given by equation 46:

$$\begin{aligned} r_i^{[l]} &= \sigma\left(U^{R[l]} h_{i-1}^{[l]} + W^{R[l]} h_i^{[l-1]} + b^{R[l]}\right) \\ d_i^{[l]} &= q_i^{[l]} \odot r_i^{[l]} \end{aligned} \tag{46}$$

Based on the result of this gate, the context signal is updated as in equation 47:

$$c_i^{[l]} = d_i^{[l]} + k_i^{[l]} \tag{47}$$

Finally, the output gate is given by equation 48:

$$\begin{aligned} s_i^{[l]} &= \sigma\left(U^{S[l]} h_{i-1}^{[l]} + W^{S[l]} h_i^{[l-1]} + b^{S[l]}\right) \\ h_i^{[l]} &= s_i^{[l]} \odot \tanh\left(c_i^{[l]}\right) \end{aligned} \tag{48}$$

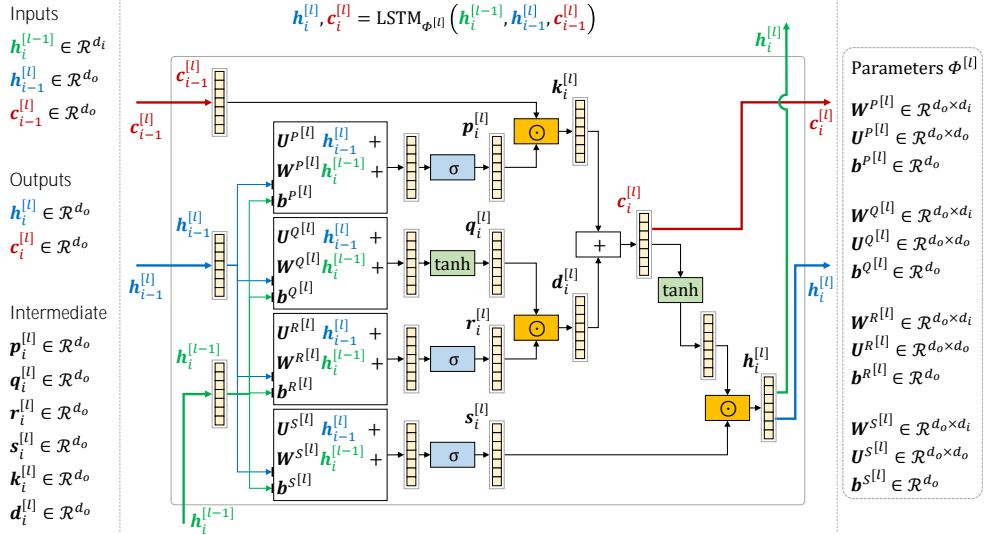

Figure 23. LSTM layer

The complete set of trainable parameters, $\Phi^{[l]}$, in an LSTM layer is shown in Figure 23 where $d_i$ and $d_o$ represent the input and output dimensionalities, respectively. The total number of trainable parameters is given by equation 49:

$$\eta_{LSTM} = 4d_o(d_o + d_i + 1) \qquad (49)$$

In practice, we set $d_i = d_o = d_e$. Replacing this in equation 49, the number of trainable parameters is given by:

$$\eta_{LSTM} = 4d_e(2d_e + 1) \qquad (50)$$

Finally, to give an idea about the depth of LSTM LMs, examples in [53] and [46] use two LSTM layers.

## 6 ADVANCED TOPICS ON TRANSFORMERS

### 6.1 Masked language modeling with BERT

As introduced in Section 3.4, masked language modeling (MLM) randomly masks some of the tokens from the input sequence and the objective is to predict them based on the remaining context [12]. Transformers were trained using the MLM objective in the seminal work [12] where the model BERT was proposed. This model was designed to pretrain deep representations from unlabeled corpora. It has a powerful structure that made it exploitable in a wide spectrum of NLP downstream tasks including Named Entity Recognition (NER), Question Answering (QA) and Natural Language Inference (NLI) through transfer learning. In order to capture text-pair representations which are important in many downstream tasks like QA and NLI, BERT was also pretrained on the Next Sentence Prediction (NSP) task: a binary classification task that aims at predicting whether a sentence B can come after a sentence A or not.

The input sequence is represented such that it satisfies the following criteria [12]:

1- A single sentence or a pair of sentences can be unambiguously fed to the model. Particularly, a special sentence separation token [SEP] is inserted at the end of each input sentence. If we need to feed a pair of sentences, the token [SEP] will be inserted at the end of each of them.
2- The input sequence always starts with a special classification token [CLS] where the objective is to generate, at the corresponding output of the model, a representation of the whole sentence which can be used in classification tasks.

Thus, the input sequence will look like $\mathcal{S} = ([CLS], w_2, w_3, ..., [SEP], ..., w_{n-1}, [SEP])$ or $\mathcal{S} = ([CLS], w_2, w_3, ..., w_{n-1}, [SEP])$ and it will be represented by one-hot vectors as $\boldsymbol{\Omega} = [\boldsymbol{\omega}_1, ..., \boldsymbol{\omega}_i, ..., \boldsymbol{\omega}_{n-1}, \boldsymbol{\omega}_n]$. The one-hot vector $\boldsymbol{\omega}_1$ will always correspond to the [CLS] token and $\boldsymbol{\omega}_n$ will correspond to the [SEP] token. Figure 24 shows the general architecture of BERT. It consists of three parts: 1) BERT backbone, 2) MLM head and 3) NSP head.

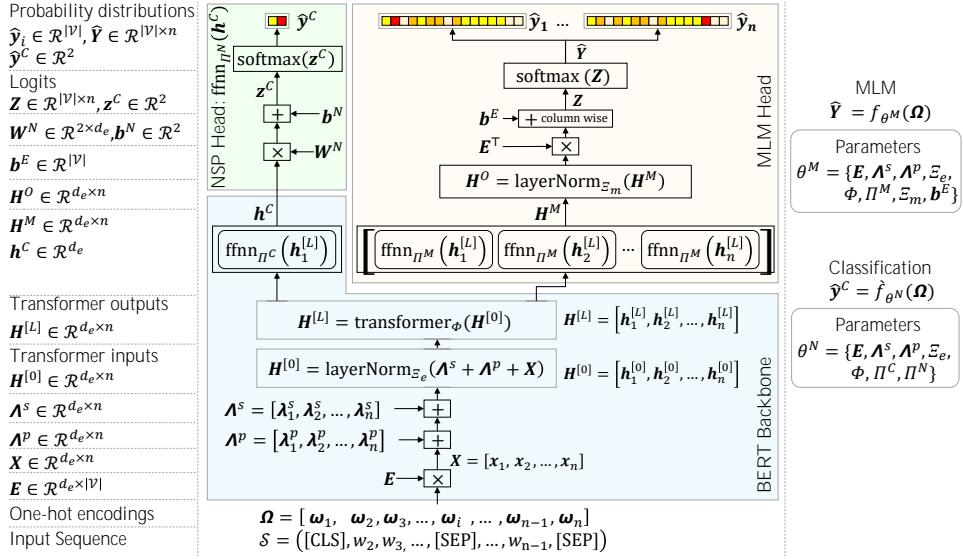

Figure 24. BERT with MLM and NSP heads

Besides the embeddings $\boldsymbol{X} \in \mathcal{R}^{d_e \times n}$ and the positional encoding $\boldsymbol{\Lambda}^p \in \mathcal{R}^{d_e \times n}$, BERT adds a third input encoding $\boldsymbol{\Lambda}^s \in \mathcal{R}^{d_e \times n}$ to differentiate between input segments i.e. sentence A and sentence B. Note that vectors $\boldsymbol{\lambda}_i^s$ are exactly the same for all tokens belonging to the same sentence. Once these three matrices $\boldsymbol{\Lambda}^s, \boldsymbol{\Lambda}^p$ and $\boldsymbol{X}$ are summed up, the resultant representation is normalized using equation 51:

$$\boldsymbol{H}^{[0]} = \text{layerNorm}_{\Xi_e}(\boldsymbol{\Lambda}^s + \boldsymbol{\Lambda}^p + \boldsymbol{X}) \qquad (51)$$

where $\Xi_e = \{\boldsymbol{\alpha}_e, \boldsymbol{\beta}_e\}$ represents the set of trainable parameters of this layerNorm function. An $L$-block transformer, whose set of trainable parameters is $\Phi = \{\Phi^{[1]}, \Phi^{[2]}, ..., \Phi^{[L]}\}$, is then applied to $\boldsymbol{H}^{[0]}$:

$$\boldsymbol{H}^{[L]} = \text{transformer}_\Phi(\boldsymbol{H}^{[0]}) \qquad (52)$$

The difference between the transformer encoder used in BERT and the transformer decoder used in GPT is only the setup of the AR/AE mask $\boldsymbol{S}$ given in equation 29. In BERT, $\boldsymbol{S} = \boldsymbol{0}_n$.

The transformer output $\boldsymbol{H}^{[L]} = [\boldsymbol{h}_1^{[L]}, \boldsymbol{h}_2^{[L]}, ..., \boldsymbol{h}_n^{[L]}]$ is then used by the MLM and the NSP heads. The MLM head takes as input the entire output sequence of the transformer $[\boldsymbol{h}_1^{[L]}, \boldsymbol{h}_2^{[L]}, ..., \boldsymbol{h}_n^{[L]}]$[7]. Vectors $\boldsymbol{h}_i^{[L]} \in \mathcal{R}^{d_e}$ are passed through a feedforward network as given in equation 53:

$$\boldsymbol{h}_i^M = \text{ffnn}_{\Pi^M}\left(\boldsymbol{h}_i^{[L]}\right) = \text{GELU}\left(\boldsymbol{W}^M \boldsymbol{h}_i^{[L]} + \boldsymbol{b}^M\right) \quad \forall i \in \{1, ..., n\} \qquad (53)$$

---

[7] In practical implementations like in KerasNLP, only columns of $\boldsymbol{H}^{[L]}$ which correspond to the masked tokens are fed as input to the MLM head (instead of passing the entire output sequence $\boldsymbol{H}^{[L]}$) in order to optimize the computation. However, we used $\boldsymbol{H}^{[L]}$ to keep the loss function formula compatible with the literature like in [62] and to make it easily comparable with the AR cost function. Note that this represents an implementation detail that doesn't change the formulas of number of trainable parameters nor the loss value.

where $\Pi^M = \{W^M, b^M\}$ represents the set of trainable parameters of this feedforward layer whose weight matrix is $W^M \in \mathcal{R}^{d_e \times d_e}$ and its bias vector is $b^M \in \mathcal{R}^{d_e}$. The resultant vectors are packaged in a matrix $H^M = [h_1^M, h_2^M, \ldots, h_n^M] \in \mathcal{R}^{d_e \times n}$. This latter is normalized using equation 54:

$$H^O = \text{layerNorm}_{\Xi_m}(H^M) \qquad (54)$$

where the set of trainable parameters $\Xi_m = \{\alpha_m, \beta_m\}$. The logits $Z = [z_1, z_2, \ldots, z_n] \in \mathcal{R}^{|\mathcal{V}| \times n}$ are then calculated using equation 55:

$$z_i = E^\top h_i^O + b^E \quad \forall i \in \{1, \ldots, n\} \qquad (55)$$

where $E \in \mathcal{R}^{d_e \times |\mathcal{V}|}$ is the embeddings matrix (we use its transpose to generate logits since we are tying input and output embeddings) and $b^E \in \mathcal{R}^{|\mathcal{V}|}$ is a bias vector. Finally, the estimated probability distributions $\hat{y}_i \in \mathcal{R}^{|\mathcal{V}|}$ are generated by simply applying the softmax function on the logits. These distributions, packaged in a matrix $\hat{Y} = [\hat{y}_1, \hat{y}_2, \ldots, \hat{y}_n] \in \mathcal{R}^{|\mathcal{V}| \times n}$, can be described as a function of the input $\Omega$ as follows:

$$\hat{Y} = f_{\theta^M}(\Omega) \qquad (56)$$

where $\theta^M = \{E, \Lambda^s, \Lambda^p, \Xi_e, \Phi, \Pi^M, \Xi_m, b^E\}$ represents the set of relevant parameters including the backbone and the MLM head of BERT as shown in Figure 24.

Concerning the NSP head, it generates a 2-compnent vector $\hat{y}^C \in \mathcal{R}^2$ based on the representation of $h_1^{[L]}$ which corresponds to the [CLS] token. The two components of $\hat{y}^C$ represent the estimated probability of having the sentence B as a valid continuation of the sentence A or not. In fact, in the backbone of BERT, the vector $h_1^{[L]}$ is first passed through a fully connected layer as given in equation 57:

$$h^C = \text{ffnn}_{\Pi^C}\left(h_1^{[L]}\right) = \tanh\left(W^C h_1^{[L]} + b^C\right) \qquad (57)$$

where $\Pi^C = \{W^C, b^C\}$ is its trainable parameters' set with $W^C \in \mathcal{R}^{d_e \times d_e}$ representing the weights matrix and $b^C \in \mathcal{R}^{d_e}$ is a bias vector. The resultant vector $h^C \in \mathcal{R}^{d_e}$ is fed to the NSP head where the probability distributions $\hat{y}^C \in \mathcal{R}^2$ are generated using the equation 58:

$$\hat{y}^C = \text{ffnn}_{\Pi^N}(h^C) = \text{softmax}(W^N h^C + b^N) \qquad (58)$$

where $\Pi^N = \{W^N, b^N\}$ is the set of trainable parameters of the latter fully connected layer with $W^N \in \mathcal{R}^{2 \times d_e}$ representing the weights matrix and $b^N \in \mathcal{R}^2$ representing a bias vector. The output of the NSP head can be represented as a function of the input $\Omega$ as follows:

$$\hat{y}^C = \hat{f}_{\theta^N}(\Omega) \qquad (59)$$

where $\theta^N = \{E, \Lambda^s, \Lambda^p, \Xi_e, \Phi, \Pi^C, \Pi^N\}$ represents the corresponding set of trainable parameters. It is worth mentioning that $h^C$ is not directly exploitable in all tasks that require sentence representations since it is trained only on the NSP task. BERT should be fine-tuned on the target downstream task in order to make this representation meaningful [12].

The number of trainable parameters of the MLM head and of the backbone of BERT are given in detail in Table 5 and Table 6, respectively. The number of trainable parameters of the NSP head is $2(d_e + 1)$. Note that this head is removed after pretraining BERT and a relevant classification head should be added to fine-tune BERT on the target downstream task. Therefore, the number of effective trainable parameters of a pretrained BERT model is given as:

$$\eta_{\text{BERT}} = d_e(|\mathcal{V}| + n + 2d_e + 8) + L\big(2Md_e(d_k + d_v) + \zeta(M(2d_k + d_v) + d_e) + 2d_e d_f + 5d_e + d_f\big) + |\mathcal{V}| \qquad (60)$$

Table 5, Number of trainable parameters of MLM Head

| | |
|---|---|
| $\text{ffnn}_{\Pi^M}$ | $\eta_8 = d_e(d_e + 1)$ |
| $\text{layerNorm}_{\Xi_m}$ | $\eta_9 = 2d_e$ |
| Embedding bias | $\eta_{10} = |\mathcal{V}|$ |
| MLM Head | $\eta_{11} = \eta_8 + \eta_9 + \eta_{10} = d_e(d_e + 3) + |\mathcal{V}|$ |

Table 6, Number of trainable parameters of BERT backbone

| | |
|---|---|
| Embedding | $\eta_1 = d_e|\mathcal{V}|$ |
| Positional encoding | $\eta_2 = d_e n$ |
| Segment encoding | $\eta_3 = 2d_e$ |
| layerNorm$_{\Xi_e}$ | $\eta_4 = 2d_e$ |
| transformer$_\Phi$ | $\eta_5 = L\big(2Md_e(d_k + d_v) + \zeta(M(2d_k + d_v) + d_e) + 2d_e d_f + 5d_e + d_f\big)$ |
| ffnn$_{\Pi^c}$ | $\eta_6 = d_e(d_e + 1)$ |
| BERT backbone | $\eta_7 = \sum_{i=1}^{6} \eta_i = d_e(|\mathcal{V}| + n + d_e + 5) + L\big(2Md_e(d_k + d_v) + \zeta(M(2d_k + d_v) + d_e) + 2d_e d_f + 5d_e + d_f\big)$ |

For practical guidelines on training and finetuning BERT [12] and other transformer-based models like GPT variants [4, 48] and RoBERTa [33], the reader is referred to [50].

## 6.2 Training transformer-based language models

As introduced in Section 4.6, AR transformer LMs are trained by minimizing the following loss function (20 repeated):

$$\mathcal{J}_{AR}(\widehat{Y}, Y) = \sum_{i=1}^{n} L_{CE}(\widehat{y}_i, y_i) = -\sum_{i=1}^{n} \log \widehat{y}_i^{(c_i)} \quad (61)$$

Recall that $\widehat{y}_i^{(j)} = p_\theta(v_j | w_{\leq i}) \; \forall j \in \{1, \ldots, |\mathcal{V}|\}$, and $y_i$ is a one-hot vector representing the true probability distribution $y_i^{(j)} = p(v_j | w_{\leq i})$ with one component $y_i^{(c_i)}$ equals to 1, where $c_i$ corresponds to the true next token index $v_{c_i} = w_{i+1}$. In order to provide a detailed formula of this loss, let $\mathcal{S}_{in} = (w_1, w_2, \ldots, w_n)$ represent the input sequence and $\mathcal{S}_{out} = (w_2, w_3, \ldots, w_{n+1})$ represent the target output sequence. The loss function aims at maximizing the estimated probability distribution of $\mathcal{S}_{out}$ i.e. $p_\Theta(\mathcal{S}_{out})$. Equation 61 can thus be rewritten as follows [62, 12]:

$$\mathcal{J}_{AR}(\widehat{Y}, Y) = -\log(p_\Theta(\mathcal{S}_{out})) = -\sum_{i=1}^{n} \log p_\Theta(w_{i+1} | w_{\leq i}) = -\sum_{i=1}^{n} \log \widehat{y}_i^{(c_i)}$$
$$= -\sum_{i=1}^{n} \log \frac{\exp(z_i^{(c_i)})}{\sum_{j=1}^{|\mathcal{V}|} \exp(z_i^{(j)})} = -\sum_{i=1}^{n} \log \frac{\exp(x_{i+1}^\top h_i^{[L]})}{\sum_{j=1}^{|\mathcal{V}|} \exp(e_j^\top h_i^{[L]})} \quad (62)$$

In masked language modelling, training aims at maximizing the estimated probability $p_\Theta(\mathcal{S}_m | \mathcal{S}_c)$ as introduced in Section 3.4 where $\mathcal{S}_c$ represents a corrupted version of the original sequence $\mathcal{S} = (w_1, w_2, \ldots, w_n)$ and $\mathcal{S}_m$ represents the masked tokens (see Section 3.4 for a concrete example). Training is based on the cross entropy between the estimated probability distribution of masked tokens $\widehat{y}_i^{(j)}$ and the corresponding true probability distributions $y_i^{(j)}$ given by equations 63 and 64, respectively:

$$\widehat{y}_i^{(j)} = p_\theta(v_j | \mathcal{S}_c) \; \forall j \in \{1, \ldots, |\mathcal{V}|\} \quad (63)$$

$$y_i^{(j)} = p(v_j | \mathcal{S}_c) = \begin{cases} 1, & v_j = w_i \\ 0, & \text{otherwise} \end{cases} \quad (64)$$

The cross entropy-based loss expresses the similarity between $\widehat{y}_i^{(j)}$ and $y_i^{(j)}$. It is given by equation 65:

$$L_{CE}(\widehat{y}_i, y_i) = -\sum_{j=1}^{|\mathcal{V}|} p(v_j | \mathcal{S}_c) \log p_\theta(v_j | \mathcal{S}_c) = -\sum_{j=1}^{|\mathcal{V}|} y_i^{(j)} \log \widehat{y}_i^{(j)} \quad (65)$$

Since $y_i^{(j)}$ is a one-hot vector, equation 65 is reduced to:

$$L_{CE}(\widehat{y}_i, y_i) = -\log \widehat{y}_i^{(c_i)} \quad (66)$$

where $c_i$ represents the index of correct class i.e. $w_i = v_{c_i}$.

Let us define a binary mask $\boldsymbol{m} = [m_1, m_2, \ldots, m_n]^\top$ as an n-dimensional vector whose components indicate the positions of the masked tokens: the component equals one if the corresponding token is masked and it is zeroed out otherwise. Then, the loss function to be minimized, $\mathcal{J}_{AE}$, is given by equation 67:

$$\mathcal{J}_{AE}(\widehat{Y}, Y) = -\log p_\Theta(\mathcal{S}_m | \mathcal{S}_c) \approx -\sum_{i=1}^{n} m_i \log p_\Theta(w_i | \mathcal{S}_c)$$

$$= -\sum_{i=1}^{n} m_i \log \hat{y}_i^{(c_i)} = -\sum_{i=1}^{n} m_i \log \frac{\exp\left(z_i^{(c_i)}\right)}{\sum_{j=1}^{|V|} \exp\left(z_i^{(j)}\right)} = -\sum_{i=1}^{n} m_i \log \frac{\exp\left(x_i^\top h_i^{[L]}\right)}{\sum_{j=1}^{|V|} \exp\left(e_j^\top h_i^{[L]}\right)} \tag{67}$$

Note that $p_\Theta(\mathcal{S}_m | \mathcal{S}_c)$ is factorized in equation 67 under the assumption that all masked tokens are independently reconstructed [12, 62]. For this reason, the factorized probability represents an approximation of the conditional probability $p_\Theta(\mathcal{S}_m | \mathcal{S}_c)$ as indicated in equation 67. Some AE LMs avoid this approximation using, for example, permutation language modeling proposed in [62]. Finally, the abovementioned loss functions can be combined with other losses when training transformer-based models. For instance, BERT is trained using the sum of the MLM loss given in equation 67 and on the NSP loss [12].

### 6.3 Why transformers

In the seminal work of Vaswani et al. [58], the main motivation underlying transformers was to overcome the sequential processing disadvantages of RNNs that were discussed in Section 4.7. Transformer operations are parallel and this enabled much efficient training. Besides this important advantage, transformers have shown higher performance upper bounds compared to other architectures. Particularly, in data-plentiful regimes, transformers outperformed other types of neural networks in a wide spectrum of applications in different fields including Natural Language Processing (NLP), Computer Vision (CV) and Time Series (TS) analysis. In order to explain the reasons underlying the latter advantage, we first introduce the concept of inductive bias of neural networks.

The inductive bias of a neural network represents the choice of the hypothesis space from which learned functions will be selected [9]. It is a restriction that we need to pose on the neural network architecture in order to make learning practically possible [9, 40]. In other words, the objective of posing the inductive bias is to improve the sample-efficiency of the model i.e. the ability of the model to exploit the available data samples and to generalize from them. The inductive bias usually reflects our prior knowledge about the task at hand [9, 10]. It is generally hard-coded in the network architecture. For instance, Convolutional Neural Networks (CNNs) have two hard-coded strong constraints on the weights: locality and weight sharing [11]. Another example is the hard-coded memory cells in RNNs [11] (see Figure 23 for the hard-coded forget, add and output gates of LSTMs).

In fact, when large amounts of training data are available, hard-coded inductive biases can be very restrictive [11]. In such regimes, relaxing the inductive bias might lead to higher performance upper bounds. In other words, this relaxation might enable a better performance / sample-efficiency trade-off.

In computer vision, it has been shown that the inductive bias of transformers is more relaxed compared to the one of CNNs [11, 14]. Only when trained on huge amounts of data, transformers outperformed the best CNN-based state-of-the-art solutions [14, 55 chapter 5]. Note that the relaxed inductive-bias of transformers is the main motivation underlying their use in computer vision applications since the parallel computing advantages are already satisfied by the state-of-the-art CNN-based solutions in this field [55 chapter 5].

In general, the advantages related to the relaxed inductive bias of transformers combined with the efficiency of their parallel architecture made them very successful in transfer learning where large models are pretrained and then reused in a multitude of downstream tasks as we described in Section 3.5. This is the case in NLP with models like GPT [4, 43, 47, 48], BERT [12], RoBERTa [33] and T5 [49], in CV with models like ViT [14], iGPT [7], CrossViT [6], PyramidViT [59] and DeiT [57], and in time series classification and prediction with models like TST [64] and PatchTST [41].

# 7 TRANSFORMERS IN COMPUTER VISION AND TIME SERIES APPLICATIONS

We explore in this section the use of transformers in computer vision and time series analysis applications. We mainly focus of the preparation of input embeddings starting from raw data.

## 7.1 Transformers in computer vision

As discussed in Section 6.3, transformers have more relaxed inductive bias compared to CNNs and this was the main motivation underlying their use in CV [11, 14, 26, 55]. For the same reason, training transformers requires large amounts of data. This explains why the impressive success of transformers in computer vision was mainly in transfer learning frameworks where large transformers are pretrained on big training sets and then reused in several applications. Several pretraining objectives were motivated by the self-supervised AR language modeling and masked language modeling tasks originally used in NLP transformers. For instance, the model iGPT [7] of OpenAI was pretrained on the auto-regressive prediction of pixels and on masked pixel prediction while the model ViT [14] of Google was tested with the masked patch prediction, among other pretraining tasks. Supervised pretraining was also possible at the presence of large labeled datasets like the JFT-300M and ImegeNet-21K. For instance, ViT [14] was pretrained on the image classification task using the aforementioned databases. Besides the pretraining objective, another important question was raised for exploiting transformers in computer vision: how to feed input images to transformers or, equivalently, how to "tokenize" the image. The model iGPT converts the input image to a 1D sequence of pixels. Thus, an input token is one pixel and the self-attention implies that each pixel attends to every other pixel [14]. Since the computational cost of transformers grows quadratically with the sequence length [29], iGPT was exploitable only with low resolution images. To overcome this problem, ViT [14] proposed a patch-based tokenization.

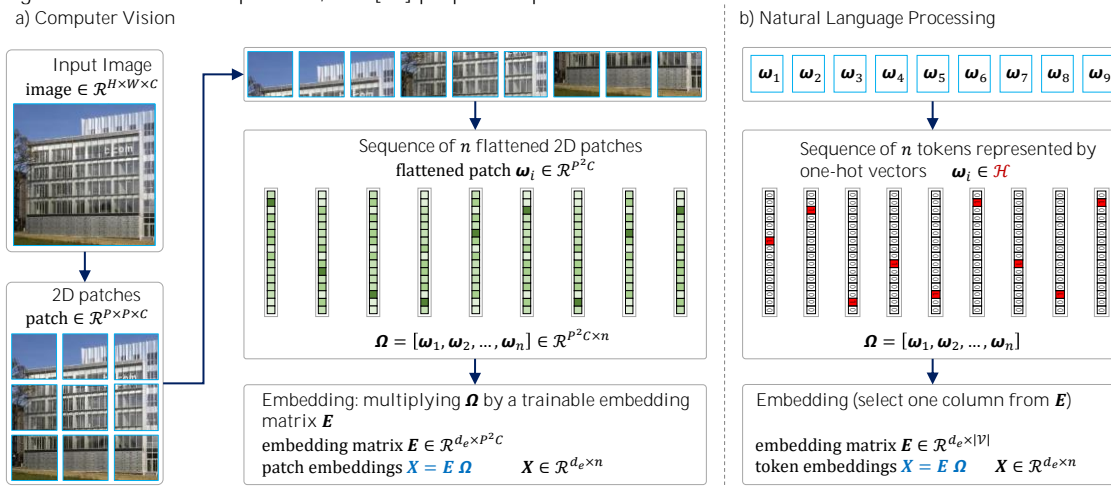

Figure 25. Preparing input embeddings in: a) Vision transformer (ViT) and b) NLP transformer

Figure 25-a shows how ViT convert the input image $I \in \mathcal{R}^{H \times W \times C}$ into $n$ patches $P_i \in \mathcal{R}^{P \times P \times C}$. These latter are then flattened to form the input sequence $\boldsymbol{\Omega} = [\boldsymbol{\omega}_1, \boldsymbol{\omega}_2, \dots, \boldsymbol{\omega}_n] \in \mathcal{R}^{P^2 C \times n}$. Patch embeddings, $X \in \mathcal{R}^{d_e \times n}$, are then obtained by multiplying the sequence by a trainable embedding matrix $E \in \mathcal{R}^{d_e \times P^2 C}$ i.e. $X = E \boldsymbol{\Omega}$. Figure 25-b shows the equivalent procedure in NLP. The resultant embeddings are added to trainable positional encodings and fed to a transformer encoder exactly as in BERT. In fact, two versions of ViT, i.e. ViT_Base with 86M parameters and ViT_Large

with 307M parameters, used the same hyper parameters of BERT while a third version, ViT_Huge with 632M parameters, was also proposed. ViT was pretrained using supervised image classification task and the version ViT_Huge showed a new state-of-the-art performance on several reference datasets. Note that ViT outperformed the CNN-based solutions only with this very large version.

ViT was also tested with self-supervised masked patch predication pretraining and showed promising results with 4% of accuracy behind supervised pretraining. For further information about vision transformers, the reader is referred to explore models like DeiT [57], PyramidViT [59] and CrossViT [6], the 5th chapter of [55] and the survey [26].

### 7.2 Transformers in time series classification, regression and forecasting

Transformer-based solutions for time series classification, regression and forecasting have shown extraordinary results in a wide spectrum of applications. For instance, in long sequence time-series forecasting, models like Informer [65], PatchTST [41] and Pyraformer [32] significantly outperformed existing methods. Transformers were also used to generate representations of multivariate time series for classification and regression problems. For instance, models like TST [64] and GTN [30] have shown very good results in the aforementioned tasks. Particularly, the performance of these two models was compared to some powerful CNN variants like ResNet [21], Inception [54] and Fully Convolutional Network [60], and showed very competitive results on several reference datasets [30, 64]. We will focus in the sequel on the TST as an example of time series transformers since its architecture is very similar to BERT.

TST represents a solution for learning representations of multivariate time series. Such representations can be used for example in classification and regression problems. Let the sequence $\boldsymbol{\Omega} = [\boldsymbol{\omega}_1, \boldsymbol{\omega}_2, \ldots, \boldsymbol{\omega}_n] \in \mathcal{R}^{C \times n}$ represent the raw time series with $n$ representing the number of time steps and $C$ representing the number of channels. The first step is to z-score the vectors $\boldsymbol{\omega}_i \ \forall \ i \in \{1, \ldots, n\}$. The resultant vectors are considered as in input "tokens". These latter are linearly projected to produce the embeddings $\boldsymbol{x}_i \in \mathcal{R}^{d_e} \ \forall \ i \in \{1, \ldots, n\}$. This is achieved using $\boldsymbol{x}_i = \boldsymbol{E}\boldsymbol{\omega}_i + \boldsymbol{b}_e \ \forall \ i \in \{1, \ldots, n\}$ where $\boldsymbol{E} \in \mathcal{R}^{d_e \times C}$ is a trainable embedding matrix and $\boldsymbol{b}_e \in \mathcal{R}^{d_e}$ is a trainable bias vector. The resultant embeddings $\boldsymbol{X} = [\boldsymbol{x}_1, \ldots, \boldsymbol{x}_n] \in \mathcal{R}^{d_e \times n}$ are added to trainable positional encodings and then fed to a transformer encoder. TST is trained using a self-supervised pretraining scheme which is similar to the MLM used in BERT. However, contrary to the classical MLM task where complete embedding vectors are masked, TST masks a proportion of each channel sequence independently as depicted in Figure 26. Masking is achieved by adding a binary noise mask to $\boldsymbol{X}$ where different rules are applied to control the length of masked and unmasked segments. Thus, training the TST is a denoising task where the Mean Square Error loss is used to compare the predictions of the masked components to their true values.

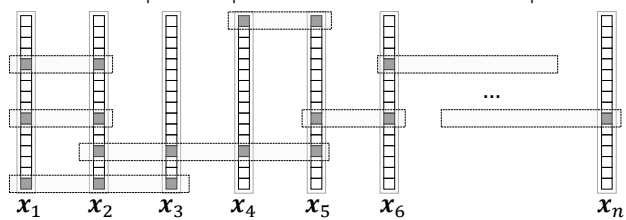

Figure 26. TST Masking scheme (reproduced from [64])

Despite the superiority of layer normalization [1] over batch normalization [24] in NLP transformers, TST uses batch normalization [24] for two reasons [64]: 1) it mitigates the effect of outliers in time series, and 2) contrary to NLP applications where the large variation in sequence length negatively affect batch normalization, the sample length variation in time series is generally much smaller. TST showed a significant performance improvement when replacing layer normalization by batch normalization [64].

ACKNOWLEDGMENTS

The authors would like to thank Salman Al-Muhammad Al-Ali, PhD student at the University of Rennes 1, for his valuable feedback that helped improving the article.

the International Conference on Learning Representations (ICLR). OpenReview.net, Toulon, France. https://openreview.net/forum?id=Bk0MRI5lg

[23] Sepp Hochreiter and Jürgen Schmidhuber. 1997. Long Short-Term Memory. Neural Comput. 9, 8, (November 1997), 1735–1780. https://doi.org/10.1162/neco.1997.9.8.1735

[24] Sergey Ioffe and Christian Szegedy. 2015. Batch normalization: accelerating deep network training by reducing internal covariate shift. In Proceedings of the 32nd International Conference on International Conference on Machine Learning - Volume 37 (ICML'15). JMLR.org, 448–456. https://dl.acm.org/doi/10.5555/3045118.3045167

[25] Dan Jurafsky. 2023. Stanford University Class CS 124: From Languages to Information. Retrieved December 08, 2023 from https://web.stanford.edu/class/cs124/

[26] Salman Khan, Muzammal Naseer, Munawar Hayat, Syed Waqas Zamir, Fahad Shahbaz Khan, and Mubarak Shah. 2022. Transformers in Vision: A Survey. ACM Comput. Surv. 54, 10s, Article 200 (January 2022), 1-41. https://doi.org/10.1145/3505244

[27] Diederik P. Kingma and Jimmy Ba. 2015. Adam: A method for stochastic optimization. In proceedings of the International Conference on Learning Representations (ICLR). San Diego, CA, USA. https://doi.org/10.48550/arXiv.1412.6980.

[28] Reinhard Kneser and Hermann Ney. 1995. Improved backing-off for M-gram language modeling. In proceedings of the International Conference on Acoustics, Speech, and Signal Processing (ICASSP). IEEE, Detroit, MI, USA, 181-184. http://dx.doi.org/10.1109/ICASSP.1995.479394

[29] Tianyang Lin, Yuxin Wang, Xiangyang Liu and Xipeng Qiu. 2022. A survey of transformers. AI Open 3, (2022), 111–132. https://doi.org/10.1016/j.aiopen.2022.10.001

[30] Minghao Liu, Shengqi Ren, Siyuan Ma, Jiahui Jiao, Yizhou Chen, Zhiguang Wang and Wei Song. 2021. Gated Transformer Networks for Multivariate Time Series Classification. arXiv:2103.14438v1. Retrieved from https://doi.org/10.48550/arXiv.2103.14438

[31] Pengfei Liu, Weizhe Yuan, Jinlan Fu, Zhengbao Jiang, Hiroaki Hayashi, and Graham Neubig. 2023. Pre-train, Prompt, and Predict: A Systematic Survey of Prompting Methods in Natural Language Processing. ACM Comput. Surv. 55, 9, Article 195 (September 2023), 35 pages. https://doi.org/10.1145/3560815

[32] Shizhan Liu, Hang Yu, Cong Liao, Jianguo Li, Weiyao Lin, Alex X. Liu and Schahram Dustdar. 2022. Pyraformer: Low-Complexity Pyramidal Attention for Long-Range Time Series Modeling and Forecasting. In proceedings of the 10th International Conference on Learning Representations (ICLR). OpenReview.net, Virtual. https://openreview.net/pdf?id=0EXmFzUn5I

[33] Yinhan Liu, Myle Ott, Naman Goyal, Jingfei Du, Mandar Joshi, Danqi Chen, Omer Levy, Mike Lewis, Luke Zettlemoyer and Veselin Stoyanov. 2020. RoBERTa: A Robustly Optimized BERT Pretraining Approach. In proceedings of the 8th International Conference on Learning Representations (ICLR). OpenReview.net. https://openreview.net/forum?id=SyxS0T4tvS

[34] Andreas Madsen, Siva Reddy, and Sarath Chandar. 2022. Post-hoc Interpretability for Neural NLP: A Survey. ACM Comput. Surv. 55, 8, Article 155 (August 2023), 42 pages. https://doi.org/10.1145/3546577

[35] Andrey A. Markov. 1913. Essai d'une recherche statistique sur le texte du roman "Eugene Onegin" illustrant la liaison des epreuve en chain (Example of a statistical investigation of the text of "Eugene Onegin" illustrating the dependence between samples in chain). Izvistia Imperatorskoi Akademii Nauk (Bulletin de l'Acad´emie Imp´eriale des Sciences de St.-P´etersbourg) 7, 153–162.

[36] Tomas Mikolov, Kai Chen, Greg Corrado, and Jeffrey Dean. 2013. Efficient estimation of word representations in vector space. In proceedings of the International Conference on Learning Representations ICLR 2013. OpenReview.net, Scottsdale, Arizona, USA. https://doi.org/10.48550/arXiv.1301.3781

[37] Tomas Mikolov, Ilya Sutskever, Kai Chen, Greg S. Corrado, Jeff Dean. 2013. Distributed representations of words and phrases and their compositionality. In proceedings of the Advances in Neural Information Processing Systems 26 (NIPS 2013). Curran Associates, Inc., Stateline, Nevada, USA. https://proceedings.neurips.cc/paper_files/paper/2013/file/9aa42b31882ec039965f3c4923ce901b-Paper.pdf

[38] Tomas Mikolov, Martin Karafiat, Lukas Burget, Jan Cernocky and Sanjeev Khudanpur. 2010. Recurrent neural network based language model. In proceedings of the Interspeech conference. ISCA, Makuhari, Chiba, Japan, 1045-1048. https://doi.org/10.21437/Interspeech.2010-343

[39] Bonan Min, Hayley Ross, Elior Sulem, Amir Pouran Ben Veyseh, Thien Huu Nguyen, Oscar Sainz, Eneko Agirre, Ilana Heintz, and Dan Roth. 2023. Recent Advances in Natural Language Processing via Large Pre-trained Language Models: A Survey. ACM Comput. Surv. 56, 2, Article 30 (February 2024), 40 pages. https://doi.org/10.1145/3605943

[40] Tom M. Mitchell. 1980. The Need for Biases in Learning Generalizations. Technical Report No. CBM-TR-117. Rutgers University, New Brunswick, NJ.

[41] Yuqi Nie, Nam H. Nguyen, Phanwadee Sinthong and Jayant Kalagnanam. 2023. A time series is worth 64 words: long-term forecasting with transformers. In proceedings of the 11th International Conference on Learning Representations (ICLR). OpenReview.net, Kigali Rwanda. https://openreview.net/pdf?id=Jbdc0vTOcol

[42] Jorge Nocedal and Stephen J. Wright. 2006. Numerical Optimization (2nd. ed.). Springer New York, NY. https://doi.org/10.1007/978-0-387-40065-5

[43] OpenAI. 2023. GPT-4 Technical Report. Technical report. OpenAI.

[44] Sinno J. Pan and Qiang Yang. 2010. A Survey on Transfer Learning. IEEE Transactions on Knowledge and Data Engineering 22, 10 (October 2010), 1345-1359. https://doi.org/10.1109/TKDE.2009.191

[45] Jeffrey Pennington, Richard Socher and Christopher Manning. 2014. GloVe: Global Vectors for Word Representation. In proceedings of the 2014 Conference on Empirical Methods in Natural Language Processing (EMNLP). Association for Computational Linguistics, Doha, Qatar. http://dx.doi.org/10.3115/v1/D14-1162

[46] Ofir Press and Lior Wolf. 2017. Using the Output Embedding to Improve Language Models. In proceedings of the 15th Conference of the European Chapter of the Association for Computational Linguistics. Association for Computational Linguistics, Valencia, Spain, 157–163. https://aclanthology.org/E17-2025

## A   APPENDICES

### A.1   Background and terminology

In this annex, definitions/descriptions of some basic terms in NLP are provided. They are mainly based on the lectures of Dan Jurafsky, Stanford University [25].

- Vocabulary: The vocabulary is the set of all distinct words that are considered in an NLP application.
- Word tokens: Word tokens are all the running words in a corpus. Note that words might not correspond to linguistic words. They might be characters, subwords or linguistic words depending on the tokenization algorithm.
- Sequence: A sequence is an ordered list of elements (it may contain multiple instances of the same element).

- Tuple: A tuple is a finite sequence. An n-tuple is a sequence of n elements, where n is a non-negative integer. Despite that sequences treated in NLP are almost always finite, the generic term sequence is more common than the term tuple or n-tuple.
- N-gram: An n-gram is a sequence of n words (unigram, bigram, trigram …). Thus an n-gram is an n-tuple of words.
- Corpus: A corpus is a computer-readable collection of text or speech (plural corpora). It can be represented by a "long" sequence of tokens $\mathcal{C} = (c_1, c_2, …, c_T)$.
- Sentence: A sentence is a sequence of word tokens. Sometimes, we loosely define a sentence as an arbitrary span of contiguous text rather than an actual linguistic sentence [12].
- Morphemes: A morpheme represents the smallest meaning-bearing unit of a language (e.g. –tion, -tive, -ible, -ing, un-, -est, -er, look).
- Subwords (word-pieces): Subwords (word-pieces) are substrings of "linguistic" words. They can be morphemes or arbitrary substrings of words.
- Word tokenization: Word tokenization is the process of separating out words from running text. Three main families of word tokenizers can be distinguished:
    - Rule-based tokenization: for example, white-space-based segmentation can be used to tokenize texts. In general, a set of rules is implemented using regular expressions to develop tokenizers of this family.
    - Character-based segmentation: in language like Chinese, it is simple and quite useful to tokenize texts on the character level.
    - Data-based tokenization: Data-based tokenization methods like Byte-Pair Encoding BPE and WordPiece are widely used in Language Models today. They learn their vocabulary from a training corpus.

Thus, depending on the tokenization algorithm, the same sentence can be segmented in different ways as shown in the example in Table 7 where the sentence "I'm looking for a job in New York." is tokenized in several ways. The "." is not considered in the 4th and 5th ways. "I'am" is considered as one token in the 1st method. "New York" is considered as one token in the 3rd, 4th and 5th ways. "looking" is considered as two tokens i.e. "look" and "##ing" in the 5th way.

Table 7, Example of word tokenization in different ways.

| Tokenizing the sentence « I'm looking for a job in New York. » | | | | | | | | | |
|---|---|---|---|---|---|---|---|---|---|
| 1 | I'm | | looking | for | a | job | in | New | York | . |
| 2 | I | 'm | looking | for | a | job | in | New | York | . |
| 3 | I | am | looking | for | a | job | in | New York | | . |
| 4 | I | am | looking | for | a | job | in | New York | | |
| 5 | I | am | look | ##ing | for | a | job | in | New York | |

When data-based methods are used in tokenization, they should be used first to create a vocabulary from a training set. In the following, we give concrete examples of different vocabularies when the three aforementioned tokenization methods are used:
1. $\mathcal{V} = \{I, you, apple, the, go, them, …, mouse\}$ where all words are "linguistic words".
2. $\mathcal{V} = \{a, b, c, d, 5, 9, …, !\}$ where all words are characters (this is useful with languages like the Chinses).
3. $\mathcal{V} = \{strong, ##er, ##ing, ##est, ##ed, z, …, mouse\}$ where a mixture of subwords (including letters /numbers) and linguistic words can be found. Such kind of vocabularies is prepared by data-based token-learning algorithms. For instance, the model BERT has $|\mathcal{V}| = 30\,000$ subwords generated by the WordPiece algorithm.
- Sentence tokenization: Sentence tokenization or (sentence segmentation) is the process of breaking up a text into individual sentences.